\documentclass[lettersize,journal]{IEEEtran}
\usepackage{amsmath,amsfonts}
\usepackage{algorithm}
\usepackage{array}
\usepackage{textcomp}
\usepackage{stfloats}
\usepackage{url}
\usepackage{verbatim}
\usepackage{graphicx}
\usepackage{cite}
\usepackage{gensymb}
\usepackage{transparent}
\usepackage{bbm}
\usepackage{textcomp}
\usepackage{stfloats}
\usepackage{amsmath,amsfonts,amssymb}
\usepackage{titlesec}
\titlespacing*{\section}
{0pt}{2.5ex plus 1ex minus .2ex}{2.3ex plus .2ex}
\titlespacing*{\subsection}
{0pt}{2.5ex plus 0.5ex minus .2ex}{1.3ex plus .2ex}
\titlespacing*{\subsubsection}
{0pt}{2.5ex plus 0.5ex minus .2ex}{1.3ex plus .2ex}
\usepackage[skip=5pt plus1pt, indent=10pt]{parskip}

\usepackage{xcolor}
\usepackage{subcaption}
\usepackage{algpseudocode}

\begin{document}

\title{Cooperative Probabilistic Trajectory Forecasting under Occlusion}

\author{Anshul Nayak, Azim Eskandarian
\thanks{Anshul Nayak is a Ph.D. student in Mechanical Engineering,
        Virginia Tech, VA 24060, USA
        {\tt\small anshulnayak@vt.edu}}
\thanks{ Azim Eskandarian is the Dean of College of Engineering
and Alice T. and William H. Goodwin Jr. Endowed Chair, 
        Virginia Commonwealth University, VA 23284, USA
        {\tt\small eskandariana@vcu.edu}}}




\maketitle

\begin{abstract}
Perception and planning under occlusion is essential for safety-critical tasks. Occlusion-aware planning often requires  communicating the  information of the occluded object to the ego agent for safe navigation.  However, communicating rich sensor information under adverse conditions during communication loss and limited bandwidth may not be always feasible. Further, in GPS denied environments and indoor navigation, localizing and sharing of occluded objects can be challenging.  To overcome this, relative pose estimation between connected agents sharing a common field of view   can be a computationally  effective way of communicating information  about surrounding objects. In this paper, we design an end-to-end network that cooperatively estimates the current states of occluded pedestrian in the reference  frame of ego agent and then predicts the trajectory with safety guarantees. Experimentally, we  show that the uncertainty-aware  trajectory prediction of occluded pedestrian by the ego agent  is almost similar to the ground truth trajectory  assuming no occlusion. The current research holds promise for uncertainty-aware navigation among multiple connected agents under  occlusion.
\end{abstract}


\section{Introduction}
\IEEEPARstart{M}{odern}  day autonomy relies on accurate detection and forecasting of other agents for navigation. Recently, end-to-end forecasting pipelines  were  developed which take raw sensor data and forecast the future intention of other agents \cite{Casas}\cite{FAF}.  Usually, the object is continually perceived by the sensor during forecasting. Yet, there are situations where the object may be partially or fully occluded, rendering detection and forecasting of such objects quite challenging \cite{Gilroy}. Recent advances in communication between multiple connected agents have been used to address detection and forecasting under occlusion \cite{Gelbal}\cite{opv2v}. In such a scenario, an  object occluded from  ego agent's field of view (FOV) is detected by other  agents such as vehicles and infrastructure  and the information is shared through vehicle-to-vehicle (v2v) or  vehicle-to-everything (v2x) communication respectively. However, effectively communicating rich sensor information from lidar and camera  across  multiple agents  is expensive and may result in communication latency \cite{Du}.   To overcome  this, only necessary information such as position, orientation and velocity of occluded object can be shared by establishing an initial pose   between agents sharing a common field of view. Subsequently, each agent can recover its own future pose based on visual odometry thus establishing continous pose recovery  between communicating agents if the initial pose is known \cite{Sridhar}\cite{Eskandarian_review}. Relative orientation is usually established through a rigid body transformation with known rotation and translation between a pair of  communicating agents \cite{zhou}. Once  relative orientation is established, critical information about occluded object as observed by other agents can be obtained by the ego agent through the established transformation in real-time  \cite{Nayak}. Although, past research have focused on both cooperative perception \cite{Merino}\cite{Kim} and relative orientation \cite{rel_ori}\cite{rel_ori_2},  to the best of our knowledge  cooperative perception for  prediction and planning in the presence of occluded objects has been  unexplored.

\begin{figure}[t]
    \centering
    \includegraphics[width=0.4\textwidth]{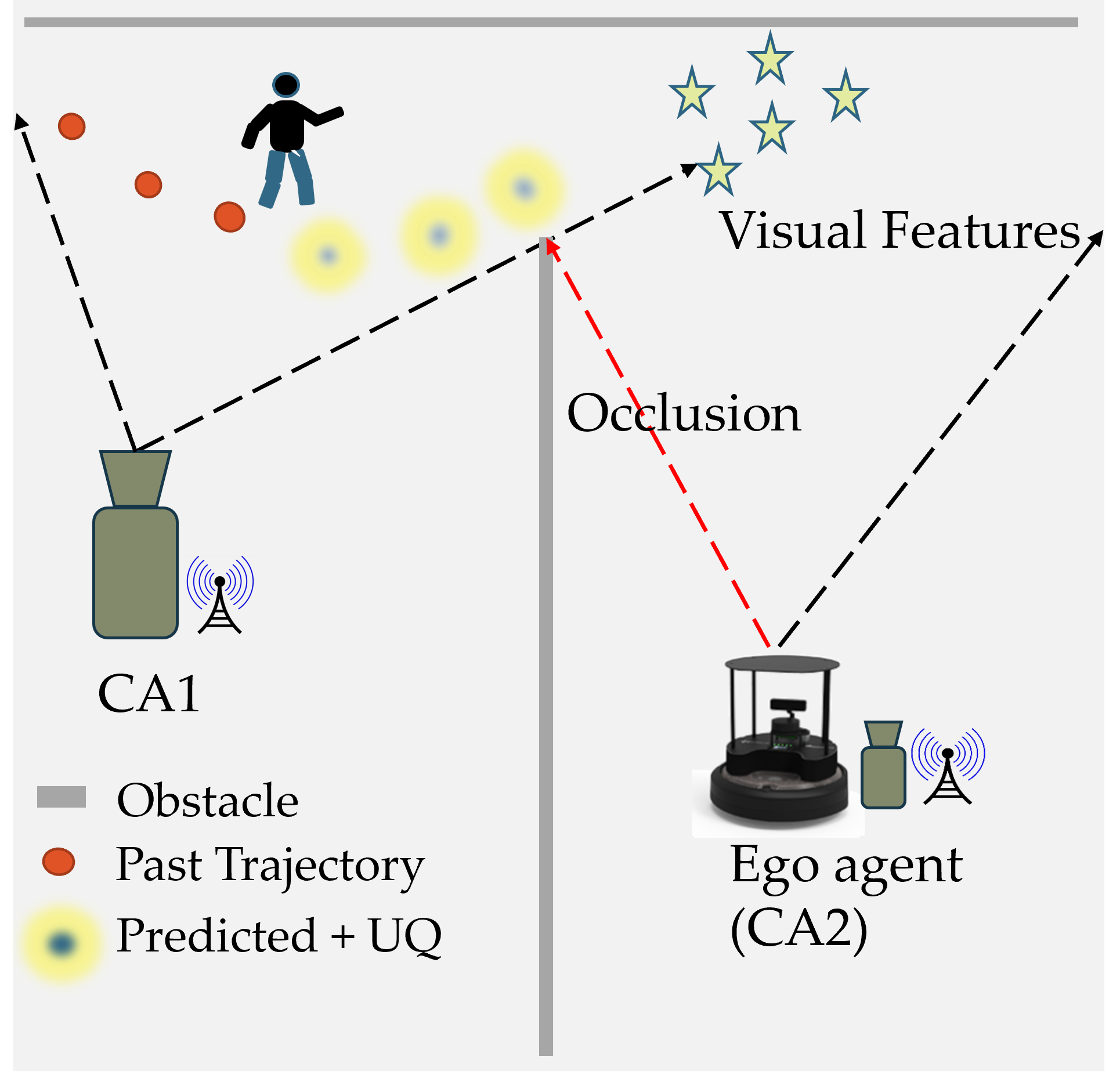}
    \caption{A Schematic of cooperative trajectory prediction under occlusion where the pedestrian is visible to Connected Agent 1 (CA1) while occluded from CA2.}
\label{fig:Coopertive_predcition}
\end{figure}

In this paper, we focus on cooperatively forecasting the trajectory of an occluded object with safety guarantees (see Figure \ref{fig:Coopertive_predcition}). In the schematic, both the connected agents CA1 and CA2 can represent infrastructures with sensors mounted in a traffic scenario \cite{multi_cam_traffic} or a multi-sensor setting \cite{multi_camera} in an indoor environment. In this work,  the connected sensors are assumed to be stationary  which can be expanded to dynamic scenarios through visual odometry as discussed earlier. Both the agents must share common some visual features such that  in order to establish relative orientation with the other agent.  However, the pedestrian is only visible to the CA1 and occluded from the view of   CA2. This makes it difficult for the ego agent to  obtain the pedestrian's  current state  for predicting the future states and ensuring safe motion planning for the ego  agent. Nonetheless, with the established cooperative perception with CA1, CA2 can  estimate the current pedestrian states as well as  predict the future states through an end-to-end prediction network \cite{MCD_Anshul}\cite{DE_Anshul}.      Note that CA2 will receive the pedestrian's location in its own frame though pose recovery and rigid body transformation and predict the future trajectory. However,  the estimated states of occluded object using relative pose estimation may not  always match the ground truth assuming no occlusion. So, deterministic prediction of future states based on estimated current states of occluded object can be overconfident. Therefore, the current prediction algorithm probabilistically predicts the future trajectory of the occluded object with safety guarantees for robustness.
We  make a few assumptions for the current research problem.\vspace{0.1cm}\
\newline\textit{Assumption 1.} CA1 and CA2 must have overlapping features at any instance for pose recovery.\vspace{0.1cm}
\newline
\textit{Assumption 2.} Both CA1 and CA2 are connected stationary sensors (depth cameras). CA2 and ego agent are the same  in the current set up.

In this paper, we would like to address the following research problems:

\textit{(Q1)}  How reliably can the ego agent  estimate the current states of the occluded pedestrian  using cooperative perception and relative pose estimation?

\textit{(Q2)} How  accurate the proposed end-to-end network's prediction is to the ground truth where CA2 can observe and predict the future trajectory of the pedestrian assuming no occlusion?

\textit{(Q3)} How  robust the proposed method is to noise  during pose recovery?

 Main Contributions of the paper are summarized as follows:

(a) We propose a 
 novel and elegant end-to-end architecture that combines   pose recovery and probabilistic trajectory forecasting 
for predicting the future states of an occluded object. To the best of our knowledge, no past literature has tried to experimentally establish the cooperative utilisation of multiple sensors in real- time trajectory prediction while handling occlusion.

(b) Experimentally, we show that the approach is robust to different long-term occlusion scenarios like partial and intermittent occlusion. This holds promise for occlusion-aware multi-agent prediction and planning in cluttered environment.

(c) Our method can be extended towards advancement of  cooperative perception especially in establishing dynamic relative pose estimation among multiple agents by combining visual odometry with initial pose recovery and help in collision avoidance during  cooperative navigation in occluded scenarios.  

\section{Literature Review}
\subsection{Relative Pose Estimation}
Usually,  relative orientation among different  agents is established  after each agent recovers its own pose based on GPS and shares the  localization  information   using   communication \cite{Fujii}. However, research shows scenarios like band width congestion or GPS denied environments can pose challenge for recovering accurate pose \cite{Hardy}. In order to address this, many relative pose estimation methods using vision \cite{Kelsey} and range \cite{zhou} measurements were developed.  Once relative orientation is established, maps can be shared and merged  to obtain information about occluded object in the surrounding. Li et.al \cite{li_hao} devised an occupancy grid map merging method where laser scans were merged  to recover relative pose between vehicles.   However, sharing and merging rich sensor information  is non-trivial and may affect safety-critical tasks in real time. In the current research, the authors propose an approach for relative pose  recovery in real time while sharing only necessary information like  position and velocity of objects in the scene using rigid body transformation 
 \cite{Sridhar}. Further, the authors also test the current  stereo vision based relative pose estimation method for robustness by injecting noise to the recovered pose.

\subsection{Trajectory Prediction}

During trajectory prediction, the future states of a dynamic object being tracked are estimated.  Traditionally, simple physics-based models such as constant velocity (CV) and constant acceleration (CA) are combined with Bayes filters to recursively estimate the future states \cite{Bayes_pred}. Most popular are Kalman Filter (KF), Extended Kalman Filter (EKF) and Particle Filter (PF) which predict future states with associated uncertainty for short horizon.  More recently, data driven approaches such as convolutional neural networks (CNN)  \cite{cnn_pred} and  Long short-term memory networks (LSTMs) \cite{lstm_pred} have been used for trajectory prediction. These models learn pattern   from distribution of trajectories and predict future trajectory based upon a sequence of past states. However, most approaches are deterministic and  
occlusion often induces uncertainty during perception and prediction of such objects.  Therefore, deterministic trajectory prediction of dynamic objects can be  over-confident and prone to error under such adverse conditions.
To improve robustness and trustworthiness, a shift towards probabilistic predictions is necessary, providing a distribution over  future states rather than single point estimates. Uncertainty-inclusive trajectory prediction   can enable safer and more reliable navigation  \cite{Kahn} \cite{Wu}. In the current research, the authors focus  upon deep probabilistic learning methods for trajectory prediction. In particular, approximate Bayesian inference method such as Monte Carlo dropout \cite{Gal} has been used to quantify uncertainty associated with  predicted states.

\subsection{Cooperative Trajectory Forecasting}

Simultaneous tracking and prediction of occluded objects has been fundamentally challenging.   Farahi et.al \cite{PKF} developed  probabilistic Kalman filter (PKF) to model trajectories using a probabilistic graph which can handle occlusion.  However, the filter is suitable at handling frame-wise partial occlusion  and may not capture completely occluded objects. Further,  Bayes filter can not handle non-linearities for long-term predictions. Alternatively, Cooperative prediction can be attained with multiple agents sharing  information with one another \cite{zhang_v2v}. 
{v2vNet \cite{v2vnet} proposes intelligently combining  information among communicating vehicles using a graph neural network to merge maps and gather information about occluded object for forecasting. However, the relative pose is assumed to be available and  broadcast as messages which may not be possible in GPS denied environments or indoor navigation. On the contrary, multi-camera multi-view tracking \cite{multi_camera}\cite{multi_camera_2}   has been implemented to track and predict occluded object's trajectory for indoor environment. The current paper leverages upon this idea of utilising multiple sensors for vision-based pose estimation in adverse scenarios for estimation and probabilistic trajectory prediction of the occluded object. The current research holds promise for occlusion-aware proactive planning under uncertainty in a cooperative manner.}

\noindent
The rest of the paper is as follows. Section III describes methods for relative pose estimation and probabilistic trajectory prediction. Section IV-A  summarizes the results for pose recovery and its sensitivity to noise. In section IV-B,  the authors experimentally show how reliably cooperative perception can be combined with  uncertainty-inclusive trajectory forecasting of occluded object under different occlusion scenarios. In Section V, conclusion alongwith future work has been discussed. 

\section{Methods}
\subsection{Relative Pose Estimation}\label{sec:rel_pose}
For connected agents, relative pose can be established between multiple agents  sharing common visual features. 
 This process consists of two fundamental steps;  feature detection and matching followed by pose recovery.

\subsubsection{ Feature Detection and Matching} 
\begin{figure}[t] 
\centering
\begin{subfigure}{.24\textwidth}
    \centering
    \includegraphics[width=0.9\linewidth]{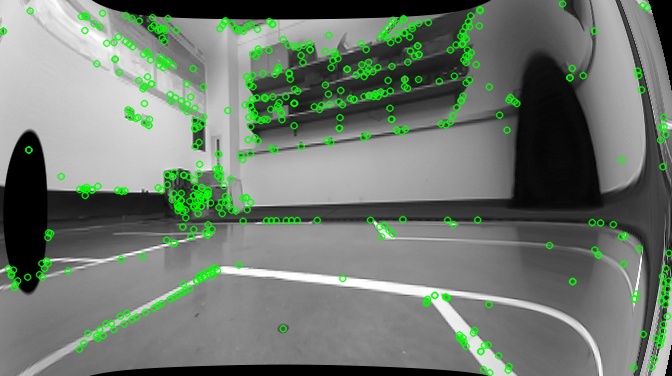}
    \caption{ CA1 Keypoints}
    \label{keypoints_1}
\end{subfigure}
    \hfill
    \begin{subfigure}{.23\textwidth}
    \centering
       \includegraphics[width = 0.95 \linewidth]{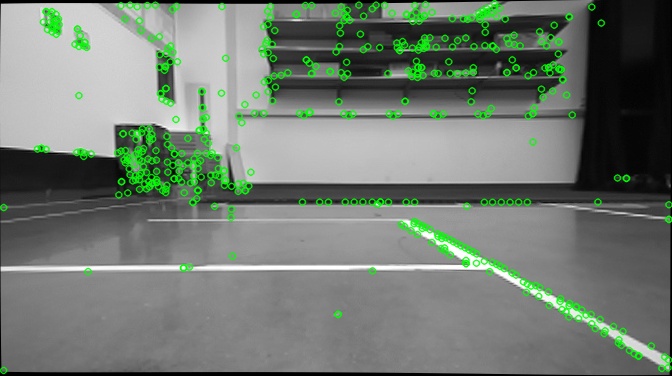}
        \caption{CA2 Keypoints  }
        \label{keypoints_2}
    \end{subfigure}
    
      \begin{subfigure}{.48\textwidth}
    \centering
       \includegraphics[width = 0.95 \linewidth]{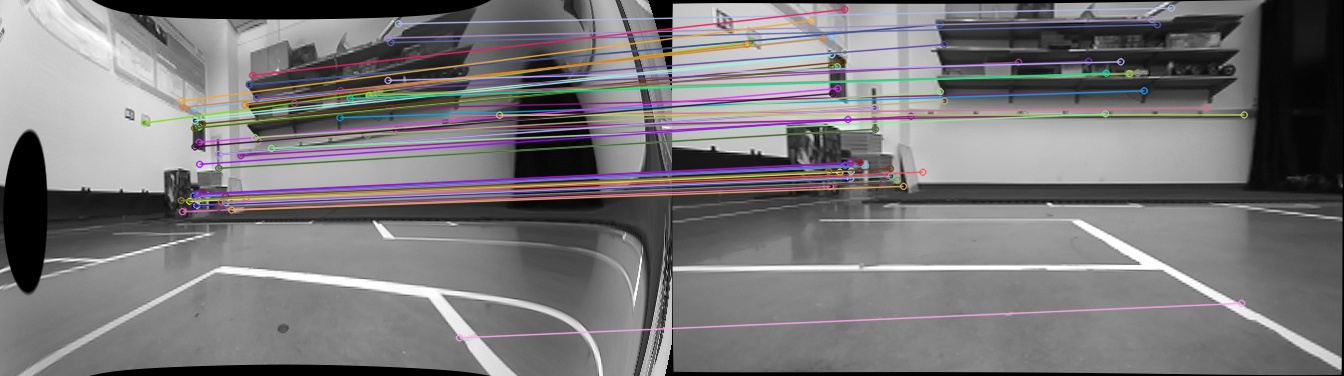}
        \caption{Feature Matching }
       \label{method:feature_matching}
    \end{subfigure}
    
    \caption{ Feature Description and Matching: (a-b) Detection of keypoints (corners, blobs, edges)   (c) Feature matching between image pair}
\end{figure}

The feature detection algorithm finds salient keypoints such as corners, edges or flat surfaces in an image (Figure \ref{keypoints_1}, \ref{keypoints_2}).  Difference of Gaussian (DoG) between images with varying Gaussian noise is used for feature detection. Each feature is then described using pixel information of a small patch around it. Descriptors can be gradient-based (like SIFT \cite{Lowe}, KAZE) which rely on  orientation of gradients in the patch or binary (like ORB \cite{ORB}, AKAZE) which generate a unique binary key for every feature. For the current study, ORB has been considered since it performs as well as SIFT with faster feature detection.   Once the features and their descriptors have been located for a pair of images, feature matching is applied based on the vectorial distance between the descriptors, as seen in (Figure \ref{method:feature_matching}). Popular matching methods include the Brute Force (BF) and FLANN-based matcher.   However, such methods match both the True Positive (TP) as well as the False Negative (FN) features. Hence, the Nearest Neighbour  Distance Ratio (NNDR) of 0.7   has been used to filter out preferred matches. 
 However, outliers may be still present with good matches which corrupt the overall pose recovery process. Therefore, RANSAC  algorithm with 1000 iterations and 99\% confidence has been applied to the matches to reject any outliers. Subsequently, the inliers are represented as homogeneous image coordinates in both the images.

\begin{figure*}[ht]
    \centering
\includegraphics[width=0.9\textwidth]{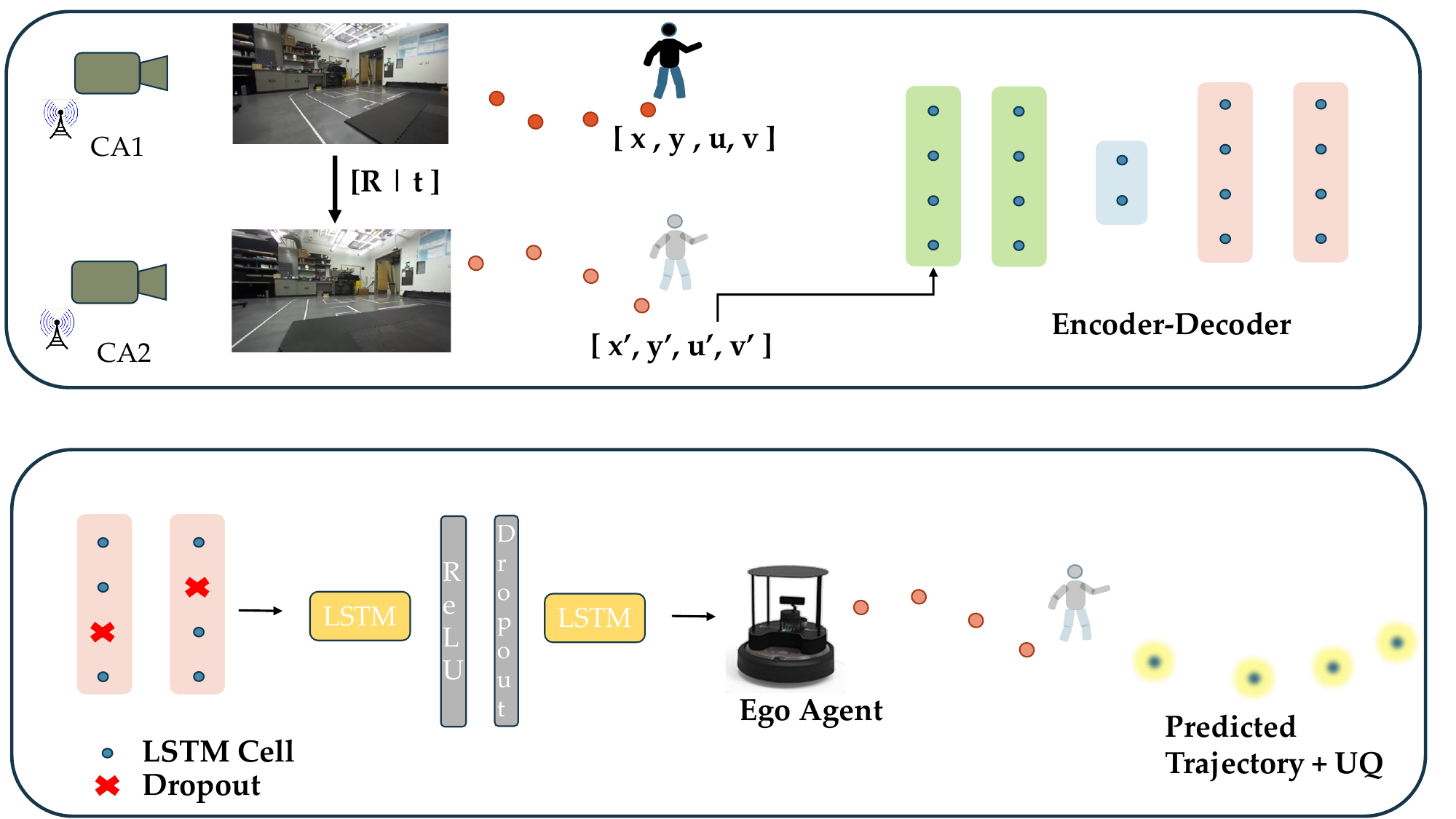}
    \caption{A schematic representation of cooperative Probabilistic Trajectory Prediction. \textbf{Top:} CA1 and CA2 recover relative pose. Occluded pedestrian's location $\mathbf{X'}$ made available to CA2 through transformation. $\mathbf{X'}$ provided to the encoder-decoder network for prediction. \textbf{Bottom:} Each LSTM layer has stochastic dropout of weights. Inference repeated multiple times to predict distribution of future states and made available to ego agent.   }
\label{fig:end_dec_model}

\end{figure*}

\subsubsection{Pose Recovery}
Any point in 3D world gets registered  in the image plane through a simple transform:
\begin{equation}
\mathrm{x} = \mathrm{P}\,\bf{X}
\end{equation}
where x is homogeneous image coordinate and $\bf{X}$ = $\mathrm{[X, Y, Z]}$ represents the point in 3D world. 
In this study, two cameras capture identical visual features from two different perspective of a scene such that the homogeneous image coordinates $\mathrm{x}$ and $\mathrm{x'}$ for each camera satisfies: 
\begin{equation}
\mathrm{x'^{T}\,F\,x = 0}
\end{equation}
Here, $\mathrm{F}$ denotes the fundamental matrix, derived via Direct Linear Transform (DLT) based on a set of 'n' matches between image pair. Given the knowledge of camera intrinsics $\mathrm{K}$, the essential matrix $\mathrm{E}$ can be computed from $\mathrm{F}$ as $\mathrm{E = K'^T FK}.$

Further, relative pose estimation between two connected agents  can be achieved through the decomposition of the Essential Matrix $\mathrm{E}$ into  rotation matrix $\mathrm{R} \in \mathbb{R}^{3\times3}$ and translation vector $\mathrm{t}\in \mathbb{R}^{3\times1}$.
\begin{equation}
  \mathrm{  E = [t]_{x}} \,\mathrm{R}
\end{equation}
The rotation matrix can be transformed into corresponding Euler angles $[\psi,\theta,\phi]$, unambiguously providing the relative orientation.
However, the essential matrix $\mathrm{E}$ is scale-invariant and the absolute distance can not be recovered. Therefore, true distance between cameras, $d_{true}$ is provided to recover the scale factor for translation vector $\mathrm{t}$.     Once the pose, $\mathrm{E = [R|t]}$ has been recovered, rigid body transformation can help facilitate in obtaining the occluded object's information   in the ego agent's frame of reference.
\begin{equation}\label{eqn:coord_transf}
    \mathrm{[X', Y', Z']}_{ego}^\mathrm{T} = \mathrm{R^{T}}\,\, \times\,\, (\mathrm{[X, Y, Z]}_{other}^\mathrm{T} - \mathrm{t})
\end{equation}

\subsection{Probabilistic Trajectory Prediction}

 Ego agent  utilizes the transformed  states $\mathbf{X'}$ based on the relative pose and predicts the future trajectory of the occluded object in a cooperative manner. As discussed,  we probabilistically predict the future trajectory with uncertainty bounds for addressing the occlusion-aware scenario rather than deterministic predictions.  For standard deterministic predictions, a simple neural network (NN) model can be trained on the pedestrian dataset and the learned model can be used for future trajectory inference. However, deterministic  future trajectory prediction can be  over-confident and prone to error. Conversely, Bayesian neural network (BNNs)
can be utilized to generate uncertainty-inclusive trustworthy future state predictions. A BNN usually has a distribution over prior weights, $\mathrm{P(\theta)}$ or activation function. The posterior distribution of the NN, $\mathrm{P}(\theta|\mathrm{D})$   can be learned from data  using Bayes Rule. 
\begin{equation}
    {\mathrm{P}}(\theta|\mathrm{D}) = \frac{\mathrm{P}(\mathrm{D}|\theta)\,{\mathrm{P}}(\theta)}{ {\mathrm{P}}(\mathrm{D})}
\end{equation}
The posterior distribution induces stochasticity during inference resulting in probabilistic prediction. For instance, provided a dataset containing the sequence of  input trajectory for a dynamic object, $X$ = {$x_{1}, x_{2}, ...,x_{T}$} and output trajectory, $Y$ = {$y_{1}, y_{2}, ...,y_{T}$}, the  distribution of its future predicted states, $y^{*}$ can be inferred by marginalizing the posterior, ${p}(\theta|\mathrm{D} = {X,Y})$ over some new input data point, $x^{*}$:
\begin{equation}
    {p}(y^{*}|x^{*},X,Y) = { \int_{\theta}^{} {p}(y^{*}|x^{*},\theta'){p}(\theta'|X,Y) \,d\theta' }
\end{equation}
$\theta'$ represents weights and  $p(\theta')$ refers to the probability of sampling from prior weight distribution. Usually, estimating the posterior  ${p}(\theta'|X,Y)$ and sampling from the distribution is quite challenging.  Some popular approximate Bayesian inference methods like Monte Carlo (MC) dropout \cite{Gal} or deep ensembles (DE) \cite{Balaji} can be used to  estimate the  distribution of predicted states without significant change to the NN architecture. The probabilistic distribution is represented  using mean and variance   assuming a Gaussian distribution over the states.  

\subsubsection{MC Dropout}
Usually, the NN is trained with dropout as a regularization   term to prevent overfitting of data. However,  MC dropout method applied during inference introduces stochastic dropout of weights  at each layer  with some probability, Bernoulli($p_{i}$). The inference process is repeated for N times for the same input, $x^{*}$ to generate a distribution of outputs $\{y_{1}^{*},y_{2}^{*},...,y_{N}^{*}\}$. As per the central limit theorem, the distribution can be approximated as Gaussian for a large N with   mean $\bar{y}^{*}$ and variance $\bar{\Sigma}y^{*}$  estimated as:
\begin{equation}
    \bar{y}^{*} = \frac{1}{N}\sum_{i=1}^{N}y^*_{i} \quad \quad   \bar{\Sigma}y^{*} = \frac{1}{N}\sum_{i=1}^{N}(y^{*}_{i} - \bar{y}^{*})^2
    \label{MCD_stats}
\end{equation}
\subsubsection{Encoder Decoder Model}
In the current research, we designed an LSTM-based encoder-decoder architecture to forecast pedestrian states over future time horizons. The encoder transforms the input trajectory sequence $\{\mathbf{{x}_{1}},\mathbf{{x}_{2}}, ..., \mathbf{{x}_{T}}  \}$ for T time steps, into an encoded space vector 'e' using a nonlinear function, i.e. e = g(x) (Figure \ref{fig:end_dec_model}).   An ablation study revealed the advantage of encoding both the position and velocity, $\mathbf{x} = \{x, y, u, v\}$, over just encoding the position. This encoded information is subsequently used by the decoder to predict future states $\{\mathbf{{x}_{T+1}}, \mathbf{{x}_{T +2}}, ..., \mathbf{{x}_{T +F}}\}$. Our system incorporates two LSTM layers each in the encoder and decoder. We employed dropout with probability $p$ and ReLU activation during training. The NN predicts the future position of pedestrians $\hat{x}$ and $\hat{y}$. During inference, MC dropout  is applied to infer the NN multiple times and  forecast a distribution for future trajectory estimates. The distribution is assumed Gaussian with mean and variance \eqref{MCD_stats}.
\vspace{0.3 cm}

\renewcommand{\thealgorithm}{: Cooperative Trajectory Forecasting}

  \begin{algorithm} 
   \caption{}
    \begin{algorithmic}
      \Function{POSE}
      {$\mathrm{x_{k}},\mathrm{x'_{k}},\mathbf{K},\mathbf{K'}$}\Comment{Where $\mathrm{x_{k}},\mathrm{x'_{k}}$ - Homogeneous Coordinates;  $\mathbf{K, K'}$ - calibration matrix}

        \For{$\mathrm{k = 1}$ to $\mathrm{N}$}
        
            \State $\mathrm{x_{k}}\, \mathrm{F}\,  \mathrm{x'_{k}} =0$ 
            \State $\mathrm{E = K'\, F\, K}$

            \State $\mathrm{E = [t]_{x}\,R}$ \Comment{Pose Recovery}

        \EndFor\\
        \Return $\mathrm{R^{3x3}}$,\,$\mathrm{t^{3x1}}$
        \EndFunction\\

    \Function{PREDICTION}{input = $\mathbf{x_{k}}$,R,t,  target = $\mathbf{[y_{k}, \Sigma_{k}^{y}]^{T}}$, epochs}\Comment{where $\mathbf{x_{k}}$ = $\mathrm{[x,y,u,v]}$}\\
        \For{$n = 1$ to $epochs$}
        \State $\mathbf{[\hat{y}_{k}, \hat{\Sigma}_{k}]} = \mathrm{Model}(\mathbf{[x_{k}] })$\Comment{Outputs}
        \State NLL = $\dfrac{\lVert \mathbf{y_{k} - \hat{y}_{k}} \rVert}{\mathbf{\hat{\Sigma}_{k}^{y}}} + \dfrac{log(\mathbf{\hat{\Sigma}_{k}^{y}})}{2} $ 
        \EndFor
    \EndFunction
\end{algorithmic}
\end{algorithm}

\section{Results}

The objective of current research is to cooperatively forecast the trajectory of an occluded object with associated uncertainty.
We perform two sets of experiments, first,  studying the reliability of pose recovery and trajectory estimation of the occluded object using the recovered pose. Secondly,  predicting the future trajectory of the occluded object with  uncertainty based on the transformed states from CA1 to  ego agent's reference.

\subsection{Trajectory Estimation of occluded object}

\begin{figure}[t] 
 \begin{center}
   \begin{minipage}{0.235\textwidth}
     \centering
     \includegraphics[width=0.9\linewidth]{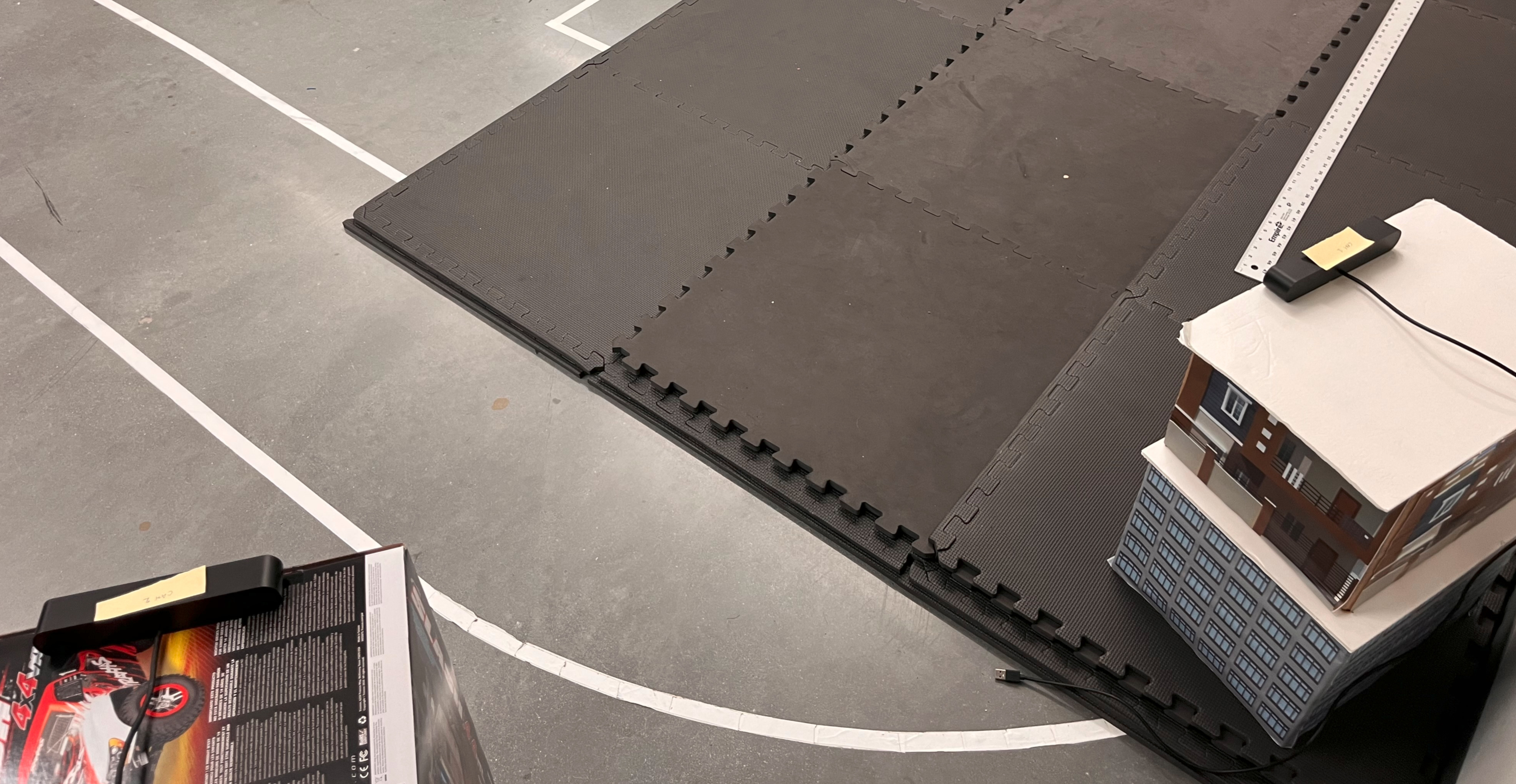}
    \subcaption{Camera Orientation}
    \label{expt:cam_orient}
    \end{minipage}
    \hfill
    \begin{minipage}{0.235\textwidth}
    \centering
       \includegraphics[width = 0.95 \linewidth]{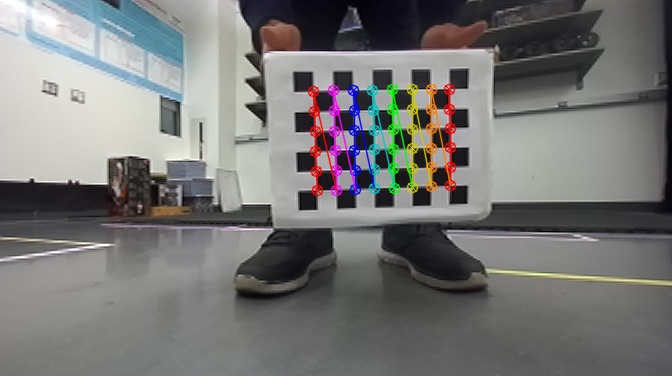}
        \subcaption{ Calibration } 
        \label{expt:calibration}
    \end{minipage}
    \begin{minipage}{0.24\textwidth}
    \centering
    \includegraphics[width=0.9\linewidth]{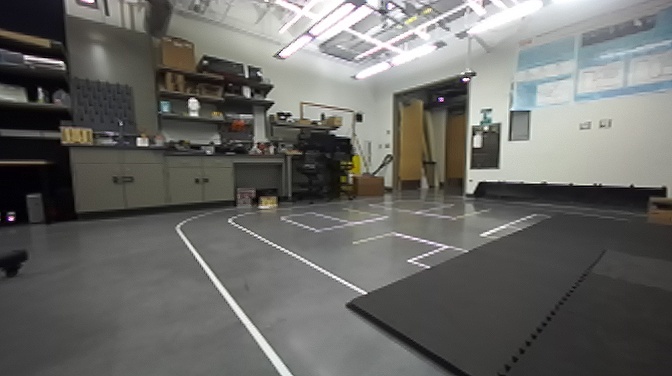}
    \subcaption{First Camera }
    \label{fig:cam_1}
    \end{minipage}
    \hfill
    \begin{minipage}{0.24\textwidth}
    \centering
       \includegraphics[width = 0.95 \linewidth]{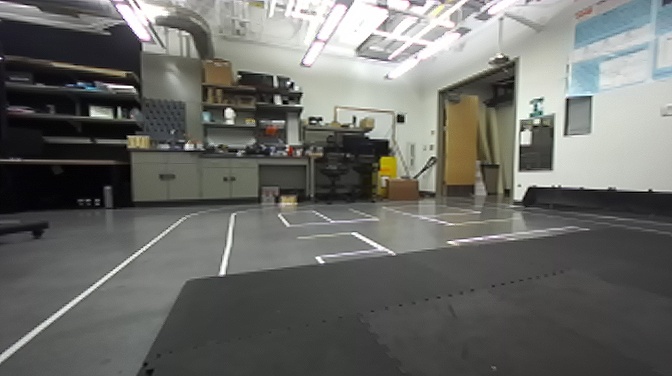}
        \subcaption{ Second Camera }
        \label{fig:cam_2}
    \end{minipage}
    \hfill
    \begin{minipage}{0.5\textwidth}
    \centering
       \includegraphics[width = 0.95 \linewidth]{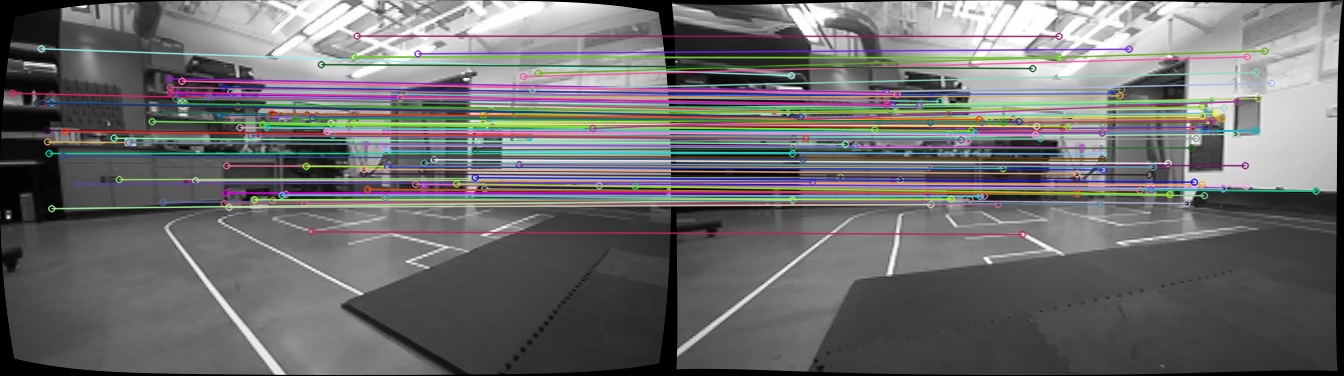}
        \subcaption{Feature Matching }
        \label{expt:feature_matching}
    \end{minipage}
   \end{center}
\label{fig:a-b-c-d-e}
    \caption{ Relative pose estimation between two cameras sharing common visual features }
\end{figure}

For establishing cooperative perception and obtaining the relative pose, we performed a simple experiment. We placed two  depth cameras with some known orientation (Figure \ref{expt:cam_orient}). We performed multi-camera  grab with software synchronisation to ensure that both the cameras capture images simultaneously with very low latency. Each camera was calibrated using a 9" x 7" checkerboard  to obtain the camera intrinsics, $\mathrm{K}$ (Figure \ref{expt:calibration}). Both the cameras share common visual features, albeit from different perspective owing to the location and orientation of each camera (Figure   \ref{fig:cam_1},\ref{fig:cam_2}). Images are transformed into gray scale  and feature description and matching are carried out based on  matched features from both images (Figure \ref{expt:feature_matching}). Relative pose, $\mathrm{[R|t]}$  between the camera pair is obtained based on the steps mentioned in 
Sec.\ref{sec:rel_pose}. The singular value decomposition of essential matrix, $\mathrm{E}$ provides four solutions for rotation matrix, $\mathrm{R}$ and translation vector, $\mathrm{t}$ \cite{Hartley} out of which only one solution is non-degenerate.

For the current experiment,  we compared experimentally obtained recovered pose  with the ground truth relative orientation   obtained directly from the internal gyroscope and accelerometer of the camera. Since, the cameras are on a  flat surface, the roll and pitch angles were negligible while the ground truth yaw orientation was $19.12^\circ$.  We tabulated the relative pose estimation for a particular orientation of cameras in Table \ref{tab:results}.

\begin{table}[h!]
\resizebox{0.45\textwidth}{!}{
    \centering
    \begin{tabular}{|c c|}
    \hline
       Ground Truth & rpy = [1.31,  -1.767,  19.12 ] \\
       \hline
         Average Estimate & rpy = [1.44,  -3.018,  21.878]\\
         \hline
         Average Feature points &  1290\\
         \hline
         Good Matches& 128 \\
         \hline
         
    \end{tabular}}
    \caption{Results for Relative Orientation}
    \label{tab:results}
\end{table}

Both ORB and SIFT based feature descriptors were able to detect significant number of feature points  along with the total good matches between the  image pair.  The feature matching time 
  to obtain the initial pose i.e essential matrix $\mathrm{E}$ was considerably  low about 87 ms. Decomposition of $\mathrm{E}$  into rotation matrix $\mathrm{R}^\mathrm{3x3}$ and translation vector $\mathrm{\hat{t}}^\mathrm{3x1}$ provides relative orientation between the cameras. $\mathrm{\hat{t}}$ is a normal vector and is scale ambiguous.    Therefore, the true distance between cameras, $d_{true}$ was estimated by measuring the depth value of the same feature point in both images.  The scale, s = $\dfrac{d_{true}}{\lVert \mathrm{\hat{t}} \rVert}$
  can be used to obtain the exact translation vector $\mathrm{t}$ = s$\mathrm{\hat{t}}$
  \eqref{eq:R_and_t}.
\begin{equation}
    \mathrm{R} = \begin{bmatrix}
       0.927  &-0.0447 &0.370\\
       0.0233 &0.997 &0.062\\
       -0.372 &-0.048 &0.926
       
    \end{bmatrix}  \quad  s\mathrm{\hat{t}}= \begin{bmatrix}
        1.163 \\
        0.066 \\
        0.040
    \end{bmatrix}
    \label{eq:R_and_t}
\end{equation}

The rotation matrix $\mathrm{R}$ can be converted to Euler angles, rpy = [1.44,  -3.018,  21.878] which closely matches the ground truth orientation obtained from imu pose data within the camera (Tab. \ref{tab:results}). 
The rotation matrix is non-degenerate as the diagonal elements of the matrix $\mathrm{R}$ are close to identity, $\mathcal{I}$. Above, we compared pose estimation for one specific orientation of cameras, subsequently, in sec.\ref{Multiple_traj}, we show how reliably cooperative perception can be used  to recover pose for multiple trajectories under varying degree of relative orientation and translation between cameras.

\begin{figure}[t] 
 \begin{center}
   \begin{minipage}{0.23\textwidth}
     \centering
     \includegraphics[width=1\linewidth]{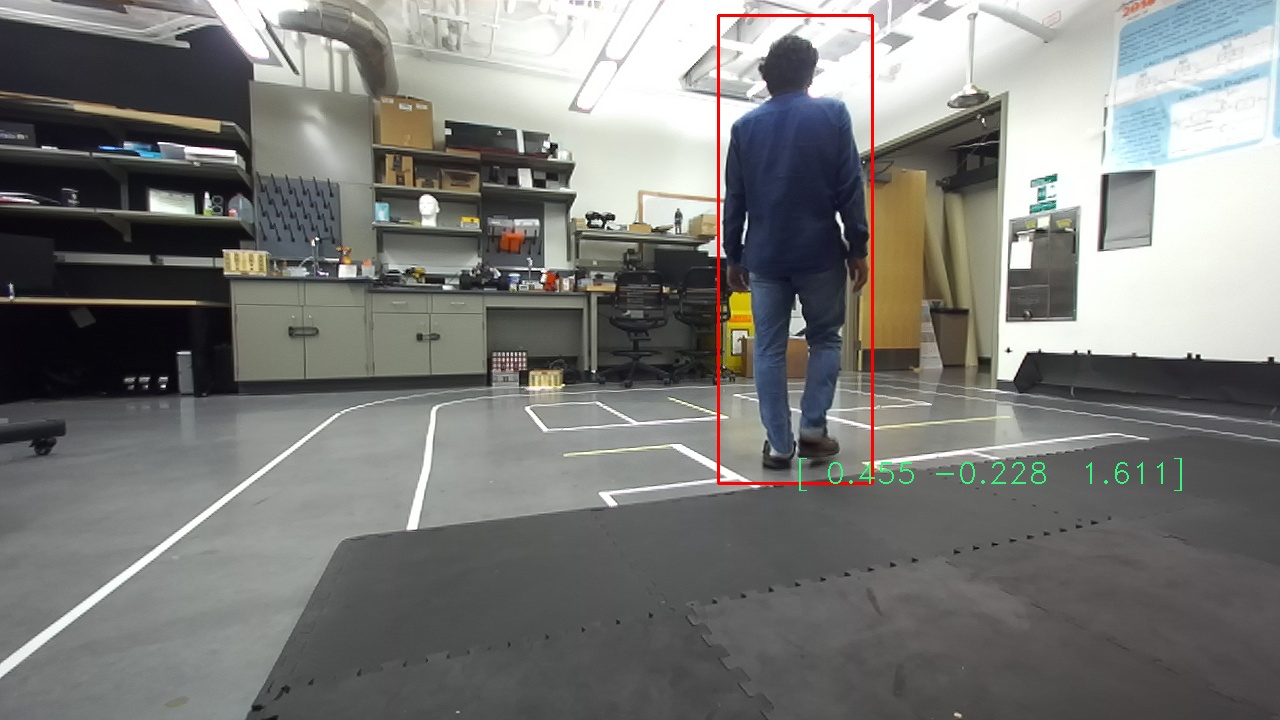}
    \subcaption{First Camera}
    \label{fig:pred_tracking_cam1}
    \end{minipage}
    \hfill
    \begin{minipage}{0.23\textwidth}
    \centering
       \includegraphics[width = 1\linewidth]{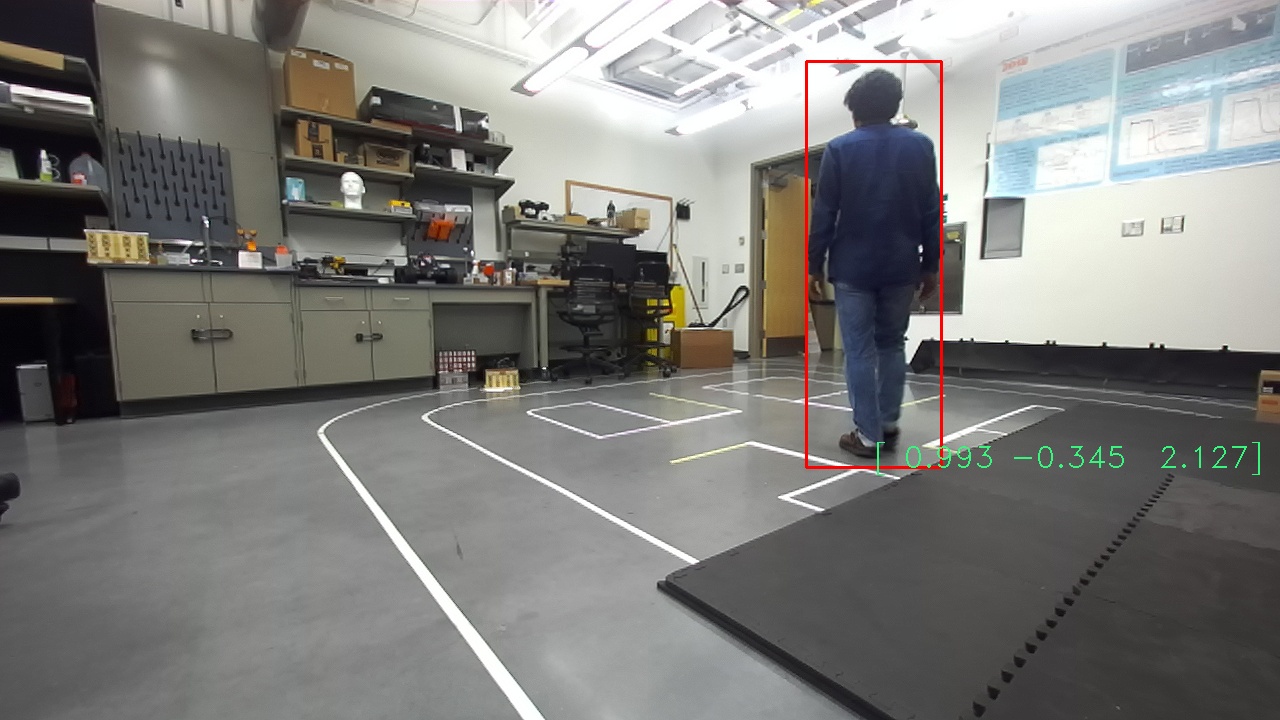}
        \subcaption{ Second Camera } 
        \label{fig:pred_tracking_cam2}
    \end{minipage}
    \end{center}
    
    \caption{Simultaneous  Object detection and tracking of the pedestrian from two cameras having some relative orientation}
    \end{figure}

Once relative orientation was established, we tracked a pedestrian simultaneously using two depth cameras  (Figure \ref{fig:pred_tracking_cam2}). The objective is to study how accurately recovered pose can be used to transform trajectory from camera 1's FOV to camera 2's reference. In reality, pedestrian shall be occluded from camera 2 and the trajectory obtained in camera 2's FOV is used as ground truth (GT) trajectory for comparison with camera 1's transformed trajectory.

\begin{figure}[h!]

        \begin{minipage}{0.15\textwidth}
     \centering
     \includegraphics[width=1\linewidth]{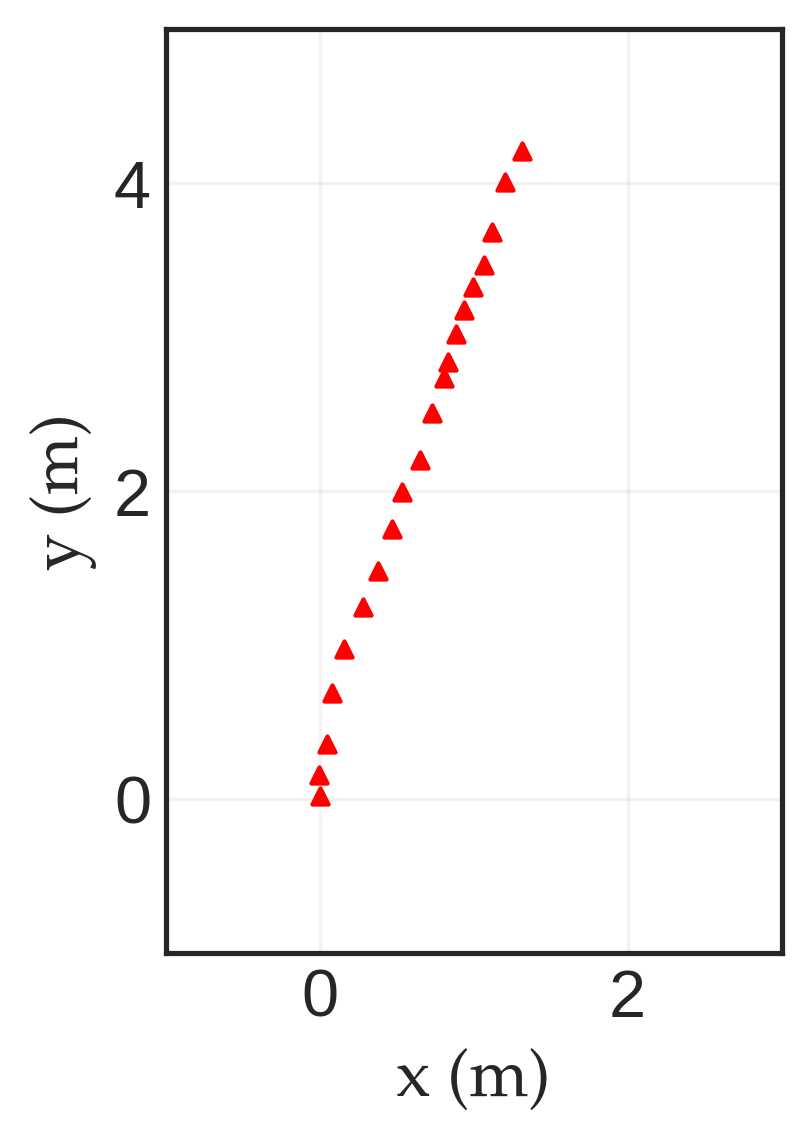}
    \subcaption{First Camera}
\label{fig:cam1_traj}
    \end{minipage}
    \hfill
    \begin{minipage}{0.15\textwidth}
     \centering
     \includegraphics[width=1\linewidth]{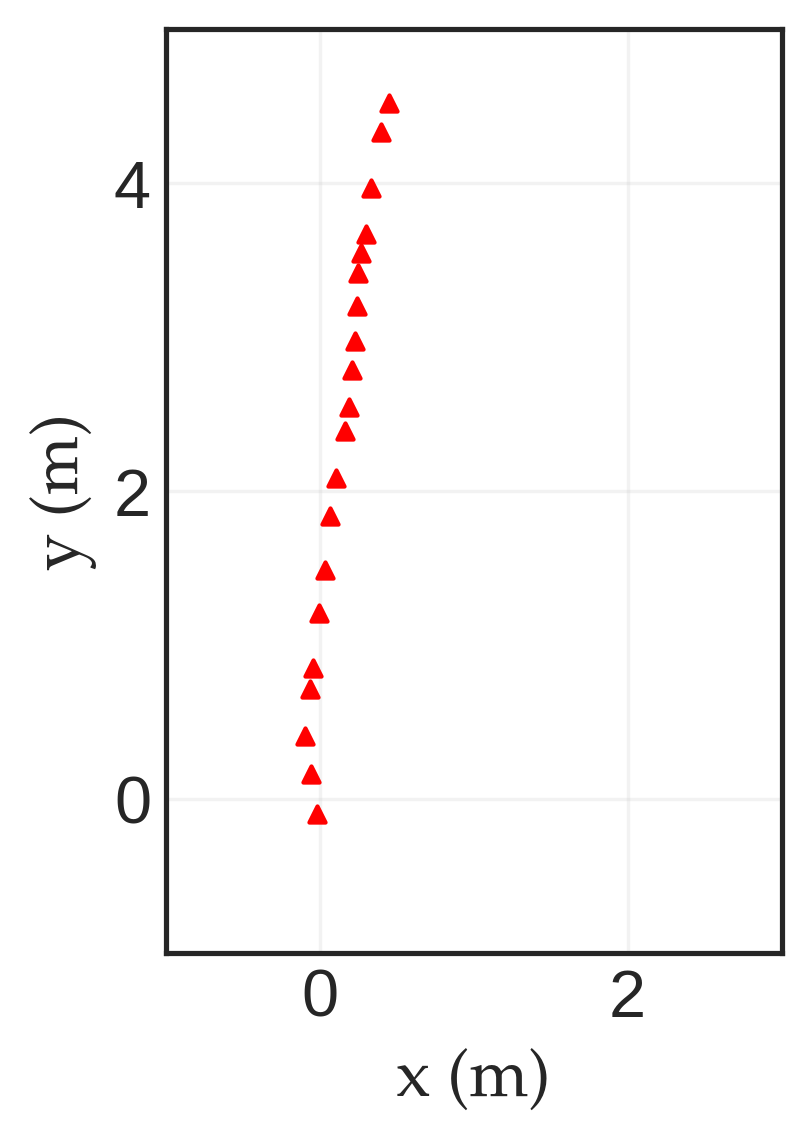}
    \subcaption{ Transformed}
    \label{fig:cam_1_transf}
    \end{minipage}
    \hfill    \begin{minipage}{0.15\textwidth}
     \centering
     \includegraphics[width=1\linewidth]{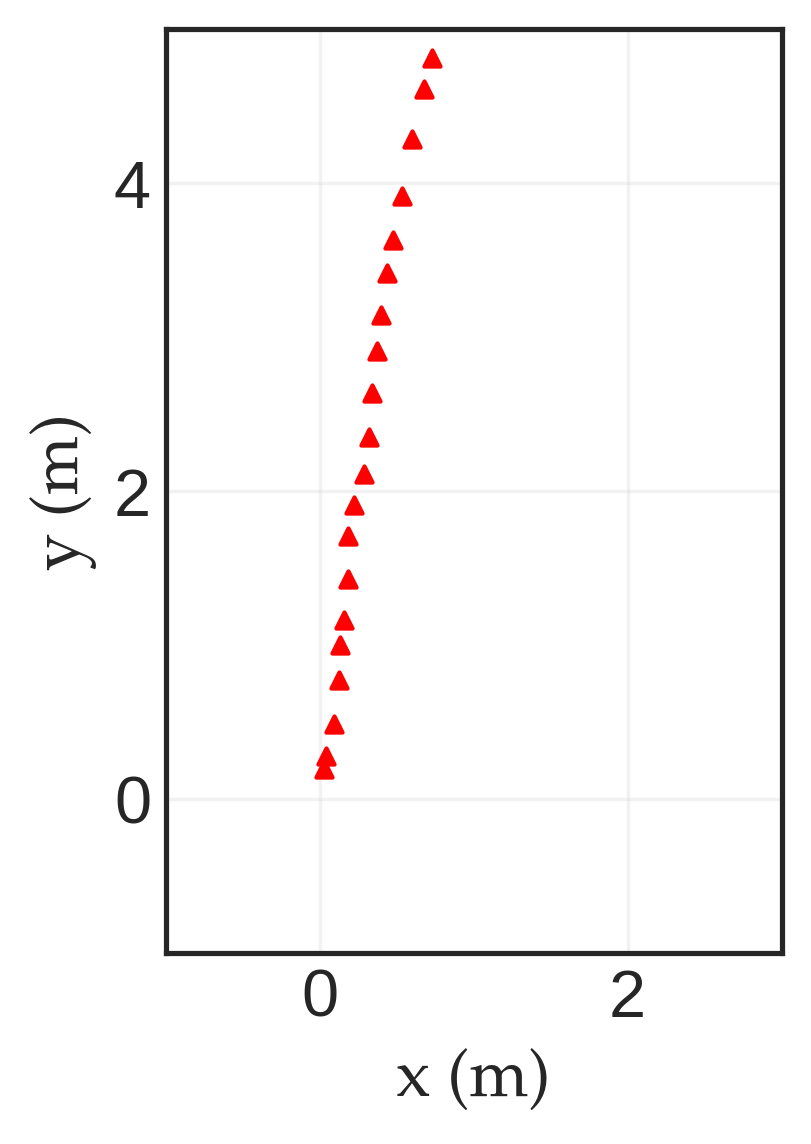}
    \subcaption{Second Camera}
    \label{fig:cam2_traj}
    \end{minipage}

    \caption{Pedestrian trajectory in the frame of reference of (a) First camera (b) Transformed trajectory of first camera using relative pose (c) Second camera}
\end{figure}

Figure \ref{fig:cam1_traj} represents the pedestrian trajectory in the first camera's reference  frame.  We compute  the transformation of pedestrian states from first camera to the second camera's reference frame (Figure \ref{fig:cam_1_transf}) using the estimated relative orientation between cameras $ \mathrm{[R|t]}$ \eqref{eq:R_and_t}. Since, the essential matrix $\mathrm{E} = \mathrm{K}_{1}^\mathrm{T} \mathrm{F} \mathrm{K}_{2}$ computes the relative orientation of the second camera with respect to first camera,  we use inverse transform \eqref{eqn:coord_transf} of rigid body coordinates to transform the pedestrian trajectory from first camera $\mathrm{[X, Y, Z]}$ to second camera's $\mathrm{[X', Y', Z']}$ reference frame.
\begin{equation}\label{eqn:coord_transf}
    \mathrm{[X', Y', Z']}^\mathrm{T} = \mathrm{R^{T}}\,\, \times\,\, (\mathrm{[X, Y, Z]}^\mathrm{T} - \mathrm{t})
\end{equation}

Figure \ref{fig:cam_1_transf} and \ref{fig:cam2_traj} represent the transformed trajectory and the ground truth trajectory in the second camera's reference frame. The plots indicate that the  average Euclidean between the trajectories was low showing the closeness of estimated trajectory of occluded pedestrian to the ground truth trajectory in second camera's reference as if assuming no occlusion.

\subsubsection{Reliability of Trajectory Estimation}\label{Multiple_traj}
\begin{figure}[h!]
    \centering
\includegraphics[width=0.5\textwidth]{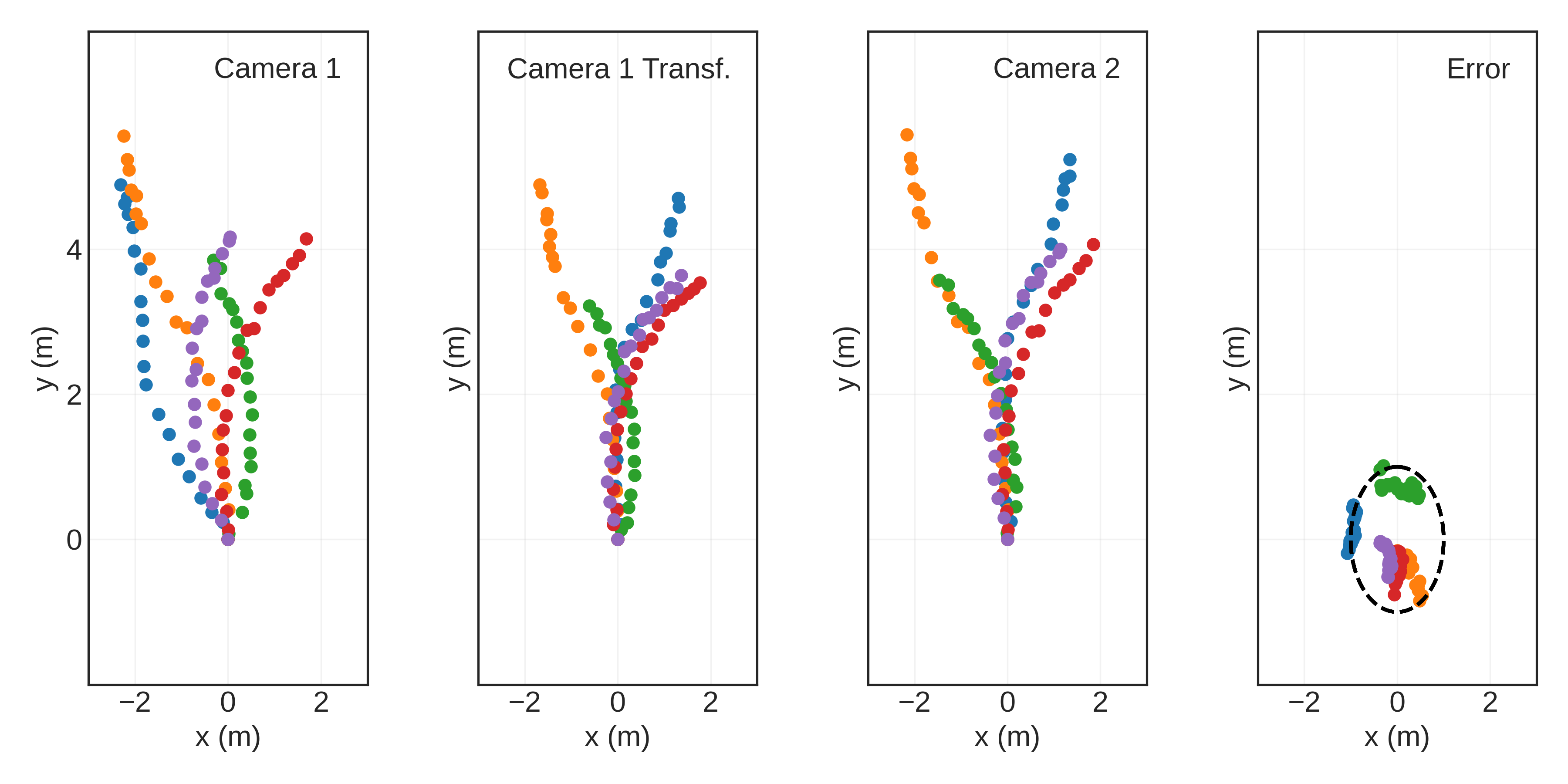}
    \caption{Cooperative estimation of multiple pedestrian trajectories with associated estimation error.}
\label{fig:mult_traj}

\end{figure}

In the previous section, we studied a single trajectory at a specific orientation of cameras. In order to  reliably estimate the state of  occluded object in camera 2's frame, we tested multiple trajectories with varying orientation of cameras.  Figure \ref{fig:mult_traj}a shows five such trajectories of tracked pedestrian in camera 1's FOV. Figure  \ref{fig:mult_traj}b shows same trajectories transformed to camera 2's frame and  Fig. \ref{fig:mult_traj}c represents trajectories as seen in camera 2's FOV. Results indicate that the error along x and y between transformed and actual trajectory in camera 2's frame of reference  is quite small (Fig. \ref{fig:mult_traj}d). Infact, the error lies within 1m of ground truth represented using dashed circle.
Overall, cooperative perception can be used to reliably estimate the relative orientation  between two CAVs and use that information to obtain the location of any occluded object in its own frame of reference.  However, as cooperative perception with pose recovery is not truly accurate as seen in Figure \ref{fig:mult_traj}, the estimation error in states can pose challenge during trajectory prediction with the transformed coordinates. Therefore, in sec.\ref{UQ_traj_pred}, we study the importance of predicting trajectory with uncertainty rather than deterministic prediction to compensate  estimation error.

 \subsubsection{Sensitivity Analysis of Relative Orientation}\label{Noise_sensitivity}

\begin{figure}[ht]
    \centering
\includegraphics[width=0.4\textwidth]{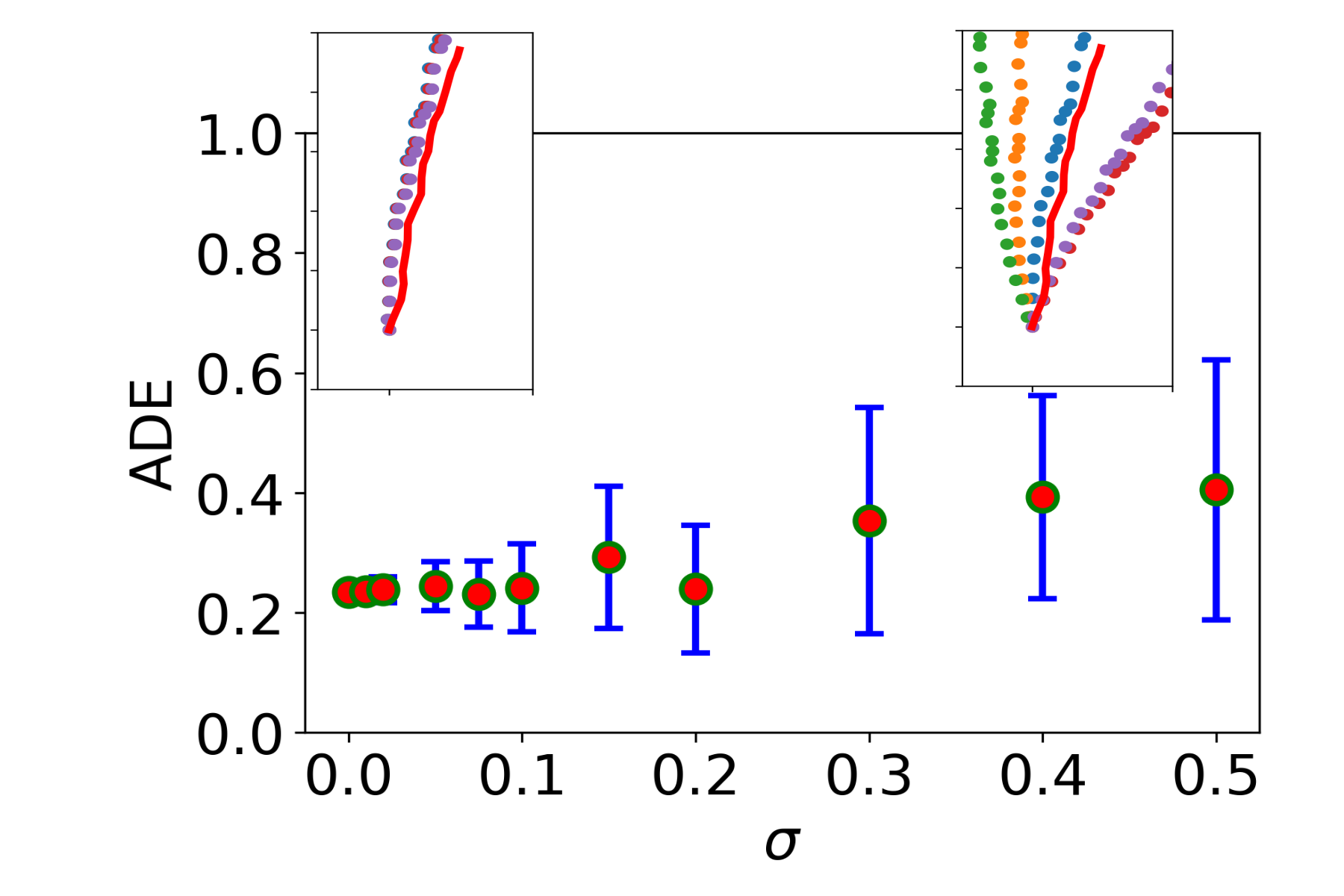}
    \caption{Variation of average displacement error with noise, $\sigma$ in Euler angles. }
\label{fig:relative_rot_error}
\end{figure}

As discussed, the vision pipeline which includes feature matching and pose recovery can be noisy and may not always result in the same  Rotation matrix $\mathrm{R}$ and translation vector $\mathrm{t}$ between a pair of images. High error in pose recovery can affect trajectory estimation of occluded object.  To simulate how noise during pose recovery affects the overall estimation of an object in another frame of reference, we conducted a sensitivity analysis. For the computed relative orientation between the cameras \eqref{eq:R_and_t}, we injected Gaussian noise $\epsilon \sim \mathcal{N}(0,\sigma^{2})$ with increasing variance to the Euler angles and translation vector $\mathrm{t}$. For the current experiments, the standard deviation, $\sigma$ was varied from 1\% to 50\% of nominal computed Euler angles as shown in Fig. \ref{fig:relative_rot_error}. We decompose the corrupted Euler angles back to a new  rotation matrix $\mathrm{R'}$ and compute the transformation of pedestrian trajectory. For every $\sigma$, the new rotation matrix $\mathrm{R'}$ was sampled 20 times and coordinate transformation \eqref{eqn:coord_transf} was applied  to the original trajectory observed in camera 1's FOV (Figure \ref{fig:cam1_traj}) to generate a  distribution of 20 transformed trajectories.
Subsequently, the average displacement error  ADE = $ \frac{1}{T}\sum_{t=t_{0}}^{t_{f}}
    ||{\hat{Y}_{(t)}- {Y}_{(t)}}|| 
   $  was computed between each transformed trajectory $\hat{Y}_{(t)}$ and the ground truth trajectory ${Y}_{(t)}$ (Fig. \ref{fig:cam2_traj})  to check the robust is trajectory transformation to noise during pose recovery $\mathrm{[R|t]}$.

Figure \ref{fig:relative_rot_error} shows the effect of corrupting the rotation matrix with Gaussian noise. The  mean and standard deviation of the ADE across 20 transformed trajectories was plotted for each $\sigma$. Results indicate that up to $\sigma = $5\%, the ADE was around 0.23  with no significant standard deviation which implies we can reliably perform coordinate transformation even if there is some noise in the recovered pose. To highlight this, we plotted a sample of 5 transformed trajectories at $\sigma = 0.02$ which shows the transformed trajectories almost overlap with little variance to the ground truth trajectory. However, at higher noise such as $\sigma$ =0.4, the transformed trajectories deviate a lot from the  ground truth (red) trajectory. No significant change was observed on the transformed trajectory when Gaussian noise  was added to the translation vector $\mathrm{t}$ without adding any noise to rotation matrix.

\subsection{Uncertainty-Inclusive Trajectory Forecasting}\label{UQ_traj_pred}

As the objective of the current research is to reliably forecast the trajectory of an occluded object  in a cooperative way, we designed the experiments accordingly. From the same orientation of cameras (Figure \ref{expt:calibration}), simultaneous object detection and tracking of the pedestrian was carried using a simple Mask R-CNN \cite{He}. The object detection module accurately classifies and tracks the pedestrian providing 3D world coordinates for position and velocity in real-time. Figure \ref{fig:pred_tracking_cam1},\ref{fig:pred_tracking_cam2} represent the tracking of pedestrian from two different perspective as seen by individual camera.   The  trajectories were collected with the camera recording at 30 frames per second for a duration of 8 seconds. The sampling time is set at 12 frames such that the camera obtains the object's position and velocity every 0.4 seconds.    Every single trajectory with
the duration of 8 seconds results in 20 $\{x, y, u, v\}$ samples, out
of which 8 samples (3.2 secs) represent past trajectory while
12 samples (4.8 secs) represent the ground truth which will be used to validate against the NN prediction.

The  NN model was trained on publicly available pedestrian datasets namely ETH\cite{Pellegrini} and UCY\cite{UCY}. End-to-end training was carried out minimising the Gaussian NLL loss with  Adam optimizer and a learning rate of 1$e^{-3}$. NN
model was trained for 150 epochs with a batch size of 32.
 The
model was compiled and fit using train and test data.
During real-time inference, only model parameters
such as trained weights and biases were considered which
makes the forward inference process computationally cheap. 

    \begin{figure}[t] 
\label{fig:cam2_traj_pred}
 \begin{center}
    \begin{minipage}{0.175\textwidth}
       \includegraphics[width = 01\linewidth]{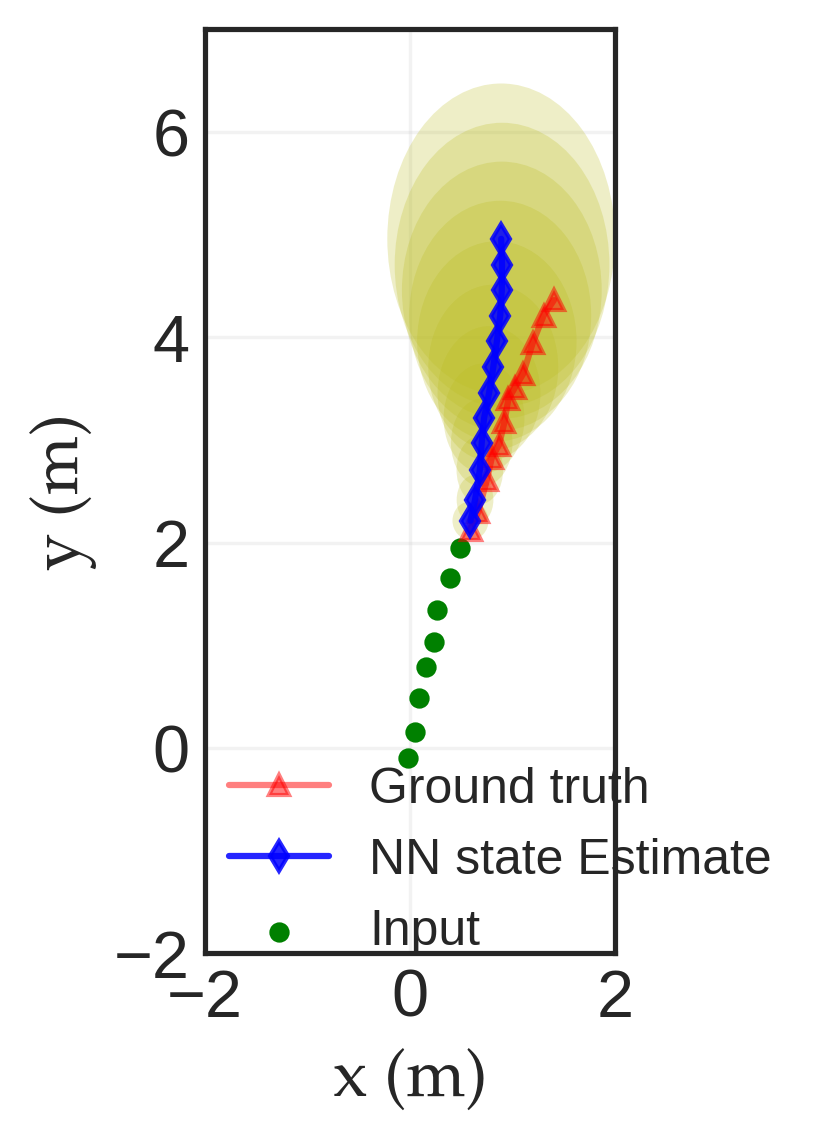} 
       \subcaption{}
    \end{minipage}
     \hfill
    \begin{minipage}{0.14\textwidth}
     \centering
     \includegraphics[width=1\linewidth]{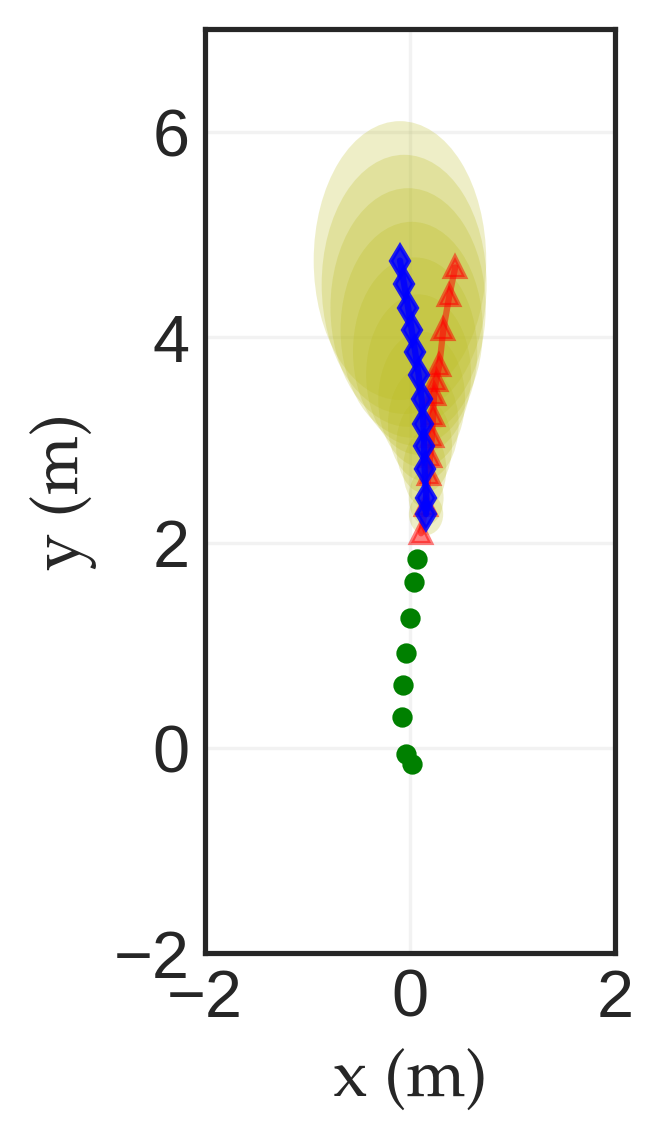}
    \subcaption{}
   \label{fig:cam1_traj_pred}
    \end{minipage}
    \hfill
        \begin{minipage}{0.14\textwidth}
     \centering
     \includegraphics[width=1\linewidth]{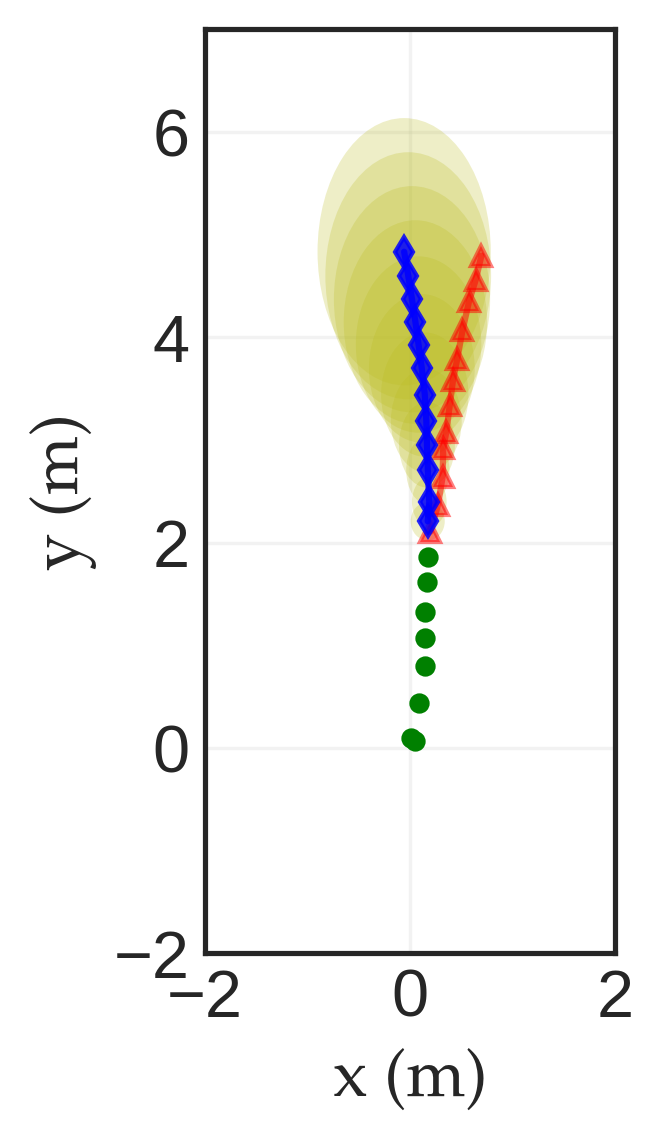}
    \subcaption{}
    \label{fig:cam2_transform_traj_pred}
    \end{minipage}
    
 \end{center} 
    
    \caption{Uncertainty-inclusive trajectory prediction for pedestrian tracking in (a) first camera
    (b) First camera transformed coordinates (c)  Second camera}  
    \label{fig:traj_pred_first_cam}
\end{figure}

In figure \ref{fig:traj_pred_first_cam}, we show 
the uncertainty-inclusive prediction for the pedestrian trajectory as observed by the first camera.  The model takes 8 input states ({\color{green} \transparent{0.5}{$\bullet$}}, green dot) to predict 12 states into future.  {\color{red} \transparent{0.65}$\blacktriangle$}  represents the actual ground truth trajectory of the pedestrian. Further, the plot shows the mean predicted path using NN({\color{blue} \transparent{0.75}{$\blacklozenge$}}, blue diamond)  alongwith the  $1\Sigma$  covariance ellipse 
 to quantify uncertainty during prediction.    The plot shows that the ADE between predicted NN state estimate and ground truth is small with the ground truth lying within the 1$\Sigma$ predictive uncertainty.

 \begin{figure*}[ht]
    \centering
\includegraphics[width=1.0\textwidth]{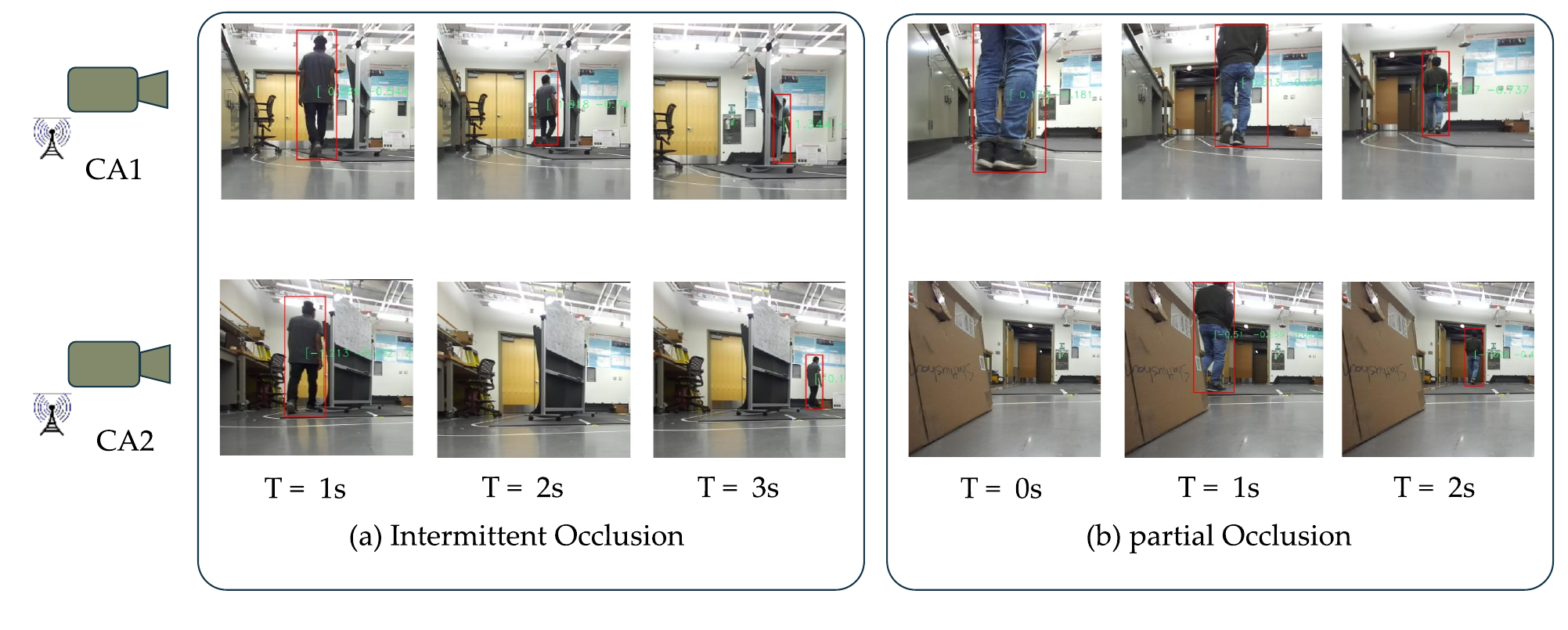}
    \caption{Scenarios representing pedestrian occluded from the FOV of camera 2. \textbf{(a) Intermittent Occlusion:} CA1 can continuously perceive the pedestrian. Pedestrian is initially visible to CA2 but occluded for some duration between T=1 to T=3 secs and again reappears. \textbf{(b) partial occlusion:}  Pedestrian is occluded initially at T=0 secs  and then continuously perceived by CA2 after T=1 secs. }
\label{fig:Occlusion_scenario}

\end{figure*}

\begin{figure}[h!]
    \centering
\includegraphics[width=0.5\textwidth]{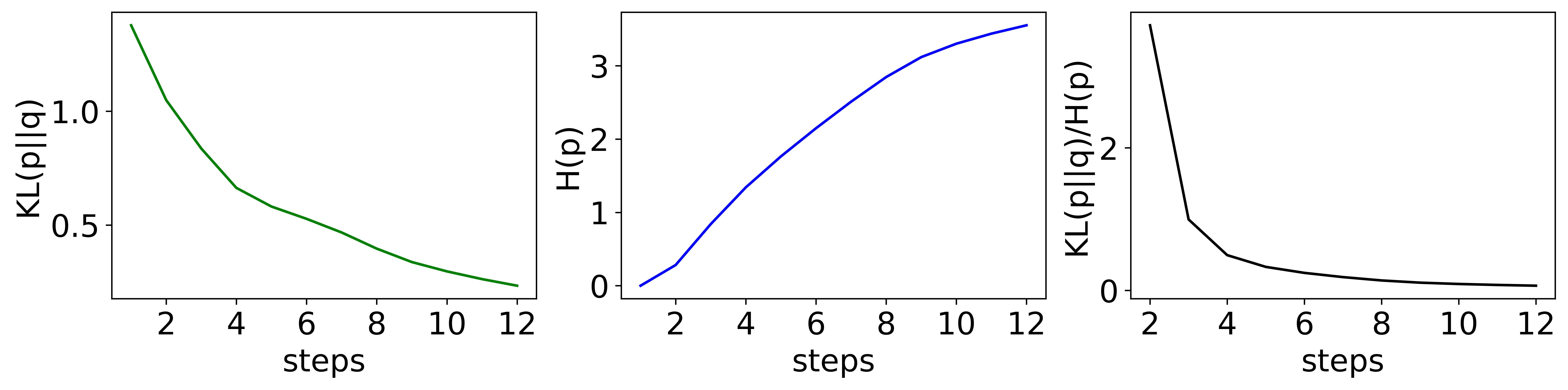}
    \caption{(a) KL divergence $\mathrm{KL}(p||q)$  (b) Shannon entropy $\mathrm{H}(p)$ and (c)  $\frac{\mathrm{KL}(p||q)}{\mathrm{H}(p)}$ of predicted distribution with future time steps.}
\label{fig:UQ_KL_div}
\end{figure}

In order to test the reliability of the cooperative perception during trajectory prediction, we transform the coordinates of the original trajectory in the first camera frame of reference which includes both the input and ground truth states  to the second camera frame using equation \eqref{eqn:coord_transf}. The transformation was applied to both  position and velocity, $\{x,y,u,v\}$ and the transformed trajectory $\{x',y',u',v'\}$  was used for prediction.  Figure \ref{fig:cam1_traj_pred}  represents the trajectory prediction with uncertainty bound for transformed trajectory. Meanwhile, Fig. \ref{fig:cam2_transform_traj_pred} represents the  prediction for the original trajectory of the pedestrian as seen by the  camera 2 assuming there is no occlusion. On inspection, the predicted mean trajectory along with the 1$\Sigma$ covariance  for future states in (Fig. \ref{fig:cam1_traj_pred}) closely matches the prediction for original trajectory (Fig.  \ref{fig:cam2_transform_traj_pred}). To quantify the dissimilarity between probabilistic predicted states, we computed the Kulback-Leibler (KL) divergence   at each  future time, $\{\mathrm{X}_{T+1}, \mathrm{X}_{T +2}, ..., \mathrm{X}_{T +F}\}$. Assuming, q $\sim \mathcal{N}_{k}(\mu_{q}, \Sigma_{q})$ and p $\sim \mathcal{N}_{k}(\mu_{p}, \Sigma_{p})$ represent the bivariate distribution of future predicted states for  Figure \ref{fig:cam1_traj_pred} and \ref{fig:cam2_transform_traj_pred} respectively at any future time $T+F$, the  KL-divergence is:
\begin{equation}
\small
    KL(p||q) = \frac{1}{2}[log\frac{|\Sigma_{q}|}{|\Sigma_{p}|} -d + tr(\Sigma_{q}^{-1}\Sigma_{p}) + (\mu_{q} - \mu_{p})^{T}\Sigma_{q}^{-1}(\mu_{q} - \mu_{p})]
\end{equation}
where d=2 represents a bivariate distribution. Similarly, the Shannon entropy which signifies the information of a true distribution is  represented as $ \mathrm{H}(p) = \frac{1}{2} log((2\pi e)^{d} det(\Sigma_{p})) $. In order to describe the distribution q  at each state, we need $\mathrm{KL}(p||q)$ bits of more information over the true distribution $\mathrm{H}(p)$. Fig.\ref{fig:UQ_KL_div}b indicates that $\mathrm{H}(p)$ increases with time implying the uncertainty in predicted distribution grows for longer prediction horizon. Meanwhile, Fig.\ref{fig:UQ_KL_div}a shows that the   KL-divergence reduces  with time implying that the predictive distributions become more and more similar between the trajectories. Further, $\frac{\mathrm{KL}(p \parallel q)}{\mathrm{H}(p)}$ which shows  how much extra bits of information would be required if we know the true distribution, p also reduces significantly. 
Overall, our results show that cooperative perception and trajectory prediction can be combined in adverse scenarios like occlusion producing almost similar results to the original ground truth predictions when there is no occlusion.

\subsubsection{Intermittent Occlusion}
In the previous section, we proposed how to reliably forecast the trajectory of an occluded dynamic object with uncertainty bound using an end-to-end network. To empirically verify the effectiveness of our proposed method, we conducted experiments to predict the path of a pedestrian in various occlusion scenarios, such as intermittent or partial occlusion (Fig. \ref{fig:Occlusion_scenario}).  In the case of intermittent occlusion, the pedestrian is initially detected by the sensor but becomes temporarily hidden from the field of view of camera 2 for a certain period before reappearing. Meanwhile, the pedestrian is continuously perceived by camera 1 (Fig. \ref{fig:Occlusion_scenario}a). We leveraged the recovered pose to convert the trajectory from camera 1 into the reference frame of camera 2, and then performed trajectory forecasting.

\begin{figure}[h!]
    \centering
\includegraphics[width=0.5\textwidth]{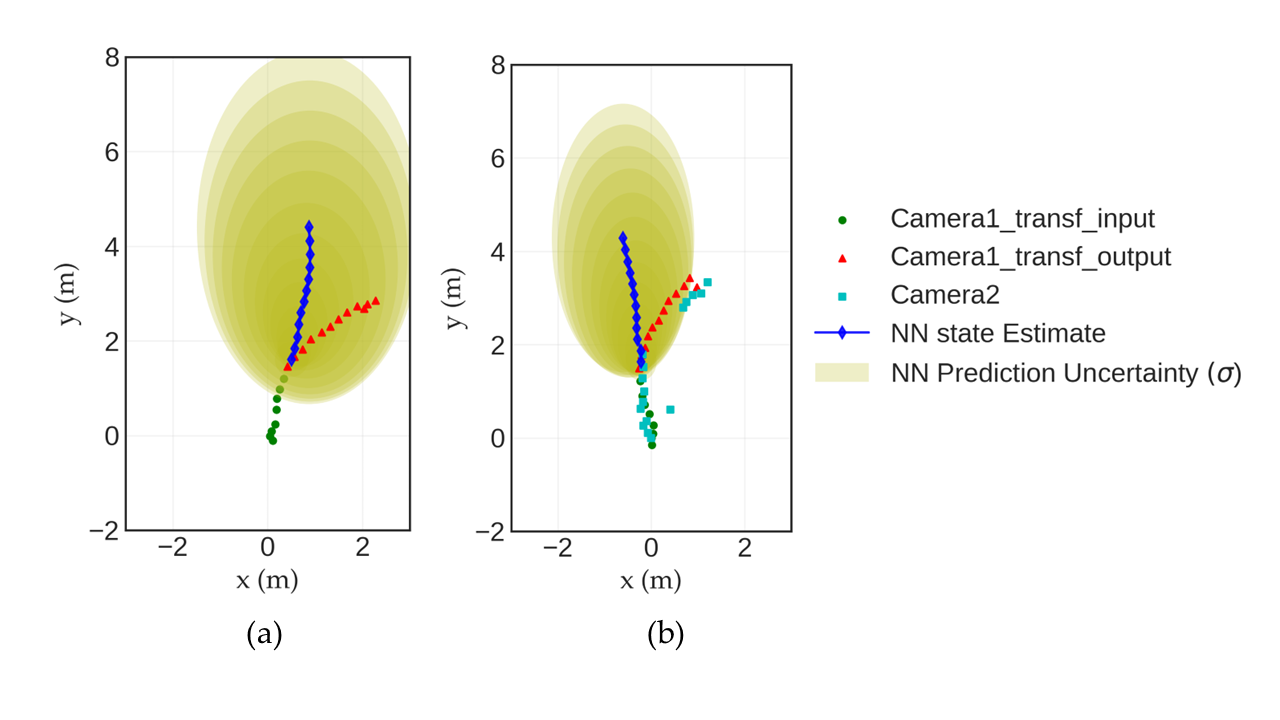}
    \caption{Uncertainty-aware prediction for trajectory in (a) Camera 1 (b) camera 1 transformed using relative pose under intermittent occlusion.}
\label{fig:intermittent_occlusion}
\end{figure}

With the established pose, trajectory estimation is accurate as the occluded pedestrian's trajectory ({\color{cyan}\transparent{0.65}${\blacksquare}$}) matches the camera 1 transformed  trajectory ({\color{green} \transparent{0.5}{$\bullet$}} + {\color{red} \transparent{0.65}$\blacktriangle$}) in Figure \ref{fig:intermittent_occlusion}b.
Meanwhile,  trajectory prediction results forecast the mean of the distribution  ({\color{blue} \transparent{0.75}{$\blacklozenge$}}) as well as the 2$\Sigma$ covariance for future states.  Fig. \ref{fig:intermittent_occlusion}b shows prediction results for transformed trajectory in camera 2's reference. Results indicate that the ground truth ({\color{red} \transparent{0.65}$\blacktriangle$}) lies within the predictive distribution for both trajectories.  However, we observed that the mean predicted path ({\color{blue} \transparent{0.75}{$\blacklozenge$}}) deviates from the actual path owing to the fact that the input trajectory ({\color{green} \transparent{0.5}{$\bullet$}}) to NN is almost straight and the NN fails to capture sudden any sudden change in direction afterwards. But nonetheless,  the actual intermittent trajectory as perceived by camera 2 ({\color{cyan}\transparent{0.65}${\blacksquare}$})  lies within the predictive distribution of the NN.

\subsubsection{Partial Occlusion}

Similar to intermittent occlusion, partial occlusion can present significant challenges in the estimation and prediction of occluded objects. In Fig. \ref{fig:Occlusion_scenario}b, we consider a scenario where a pedestrian is initially entirely obscured from the field of view of camera 2, only to reappear after a certain duration. Such situations are particularly demanding and frequently encountered in the context of traffic scenarios or indoor navigation, where pedestrians can abruptly emerge from occluded regions.

\begin{figure}[h!]
    \centering
\includegraphics[width=0.5\textwidth]{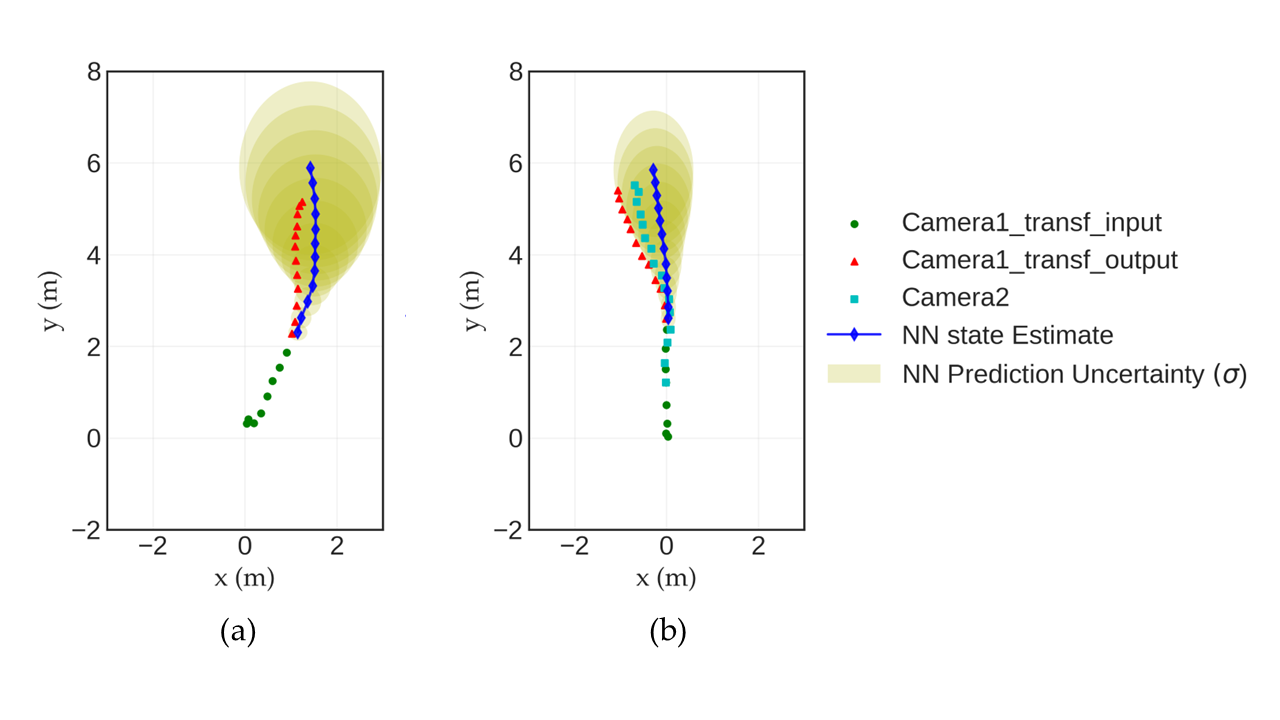}
    \caption{Uncertainty-aware prediction for trajectory in (a) Camera 1 (b) camera 1 transformed using relative pose under partial occlusion.}
\label{fig:partial_occlusion}
\end{figure}

Fig. \ref{fig:partial_occlusion}a,\ref{fig:partial_occlusion}b show  prediction results with 1$\Sigma$ covariance bound for camera 1 and camera 1 transformed trajectory.   8 input states ({\color{green} \transparent{0.5}{$\bullet$}}, green dot) of Camera 1 transformed trajectory is used to predict 12 states representing the mean NN predicted path ({\color{blue} \transparent{0.75}{$\blacklozenge$}}).  Remarkably, the ground truth ({\color{red} \transparent{0.65}$\blacktriangle$}) lies within the 1$\Sigma$ predictive distribution. More importantly, the partial trajectory ({\color{cyan}\transparent{0.65}${\blacksquare}$}) as perceived by camera 2 very closely aligns with the transformed trajectory and has a small average displacement error (ADE) with respect to the predicted path. Overall, both the experiments show how cooperative perception with pose recovery can be leveraged to probabilistically forecast the future trajectory of an object under occlusion. 

\section{Conclusion}
In this study, we have highlighted the critical role of cooperative perception in addressing challenging scenarios, such as occlusion, by integrating relative pose estimation and trajectory prediction. Our approach begins by leveraging cooperative perception between two cameras with shared visual features to accurately determine the relative pose between them. Subsequently, we conducted a series of experiments that demonstrated the robustness of estimating trajectories in the ego camera's reference frame using the recovered pose information. Our results indicate that the transformed trajectories closely match the trajectories recorded by the ego agent assuming there is no occlusion.
With the estimated trajectory, we  perform trajectory forecasting while accounting for uncertainties. The results illustrated that our probabilistic predictions for future states closely aligned with ground truth observations. Furthermore, our end-to-end prediction network can collaboratively predict trajectories under various occlusion scenarios.

While our paper presents a promising methodology, it is essential to acknowledge its limitations, notably the reliance on static cameras in our experiments.  As a future scope, researchers can aim to employ visual odometry in conjunction with initial pose estimation to reliably establish cooperative perception among multiple dynamic agents. This will allow for occlusion-aware estimation and prediction for multi-agent systems in dynamic scenarios.



	\begin{IEEEbiography}[{\includegraphics[width = 1 in, height = 1.25 in, clip, keepaspectratio]{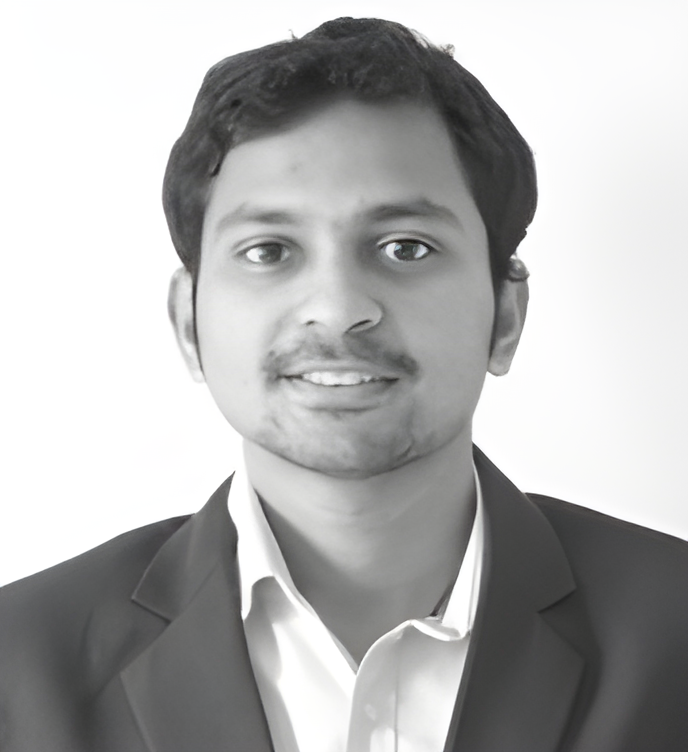}}]{Anshul Nayak}
	
		received his B.Tech  in mechanical engineering from NIT, Rourkela, India. He completed his Master's degree in Mechanical engineering at Virginia Tech and is currently pursuing his Ph.D at the  Autonomous Systems and Intelligent Machines (ASIM) lab at the same university. His research interests include uncertainty-aware  prediction and planning.
		
	\end{IEEEbiography}


\begin{IEEEbiography}[{\includegraphics[width = 1 in, height = 1.25 in, clip, keepaspectratio]{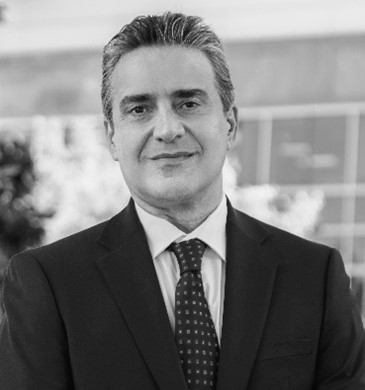}}]{Dr. Azim Eskandarian}
	
 has been Dean of the College of Engineering and Alice T. and William H. Goodwin Jr. Endowed Chair/Professor at Virginia Commonwealth University, Richmond, VA, since August 2023. Before that, he was a Professor and Head of the Mechanical Engineering Department at Virginia Tech since August 2015. He became the Nicholas and Rebecca Des Champs chair professor in April 2018 and a joint courtesy Electrical and Computer Engineering professor in 2021. He established the Autonomous Systems and Intelligent Machines laboratory at Virginia Tech, where he has conducted pioneering research in autonomous vehicles, human/driver cognition and vehicle interface, advanced driver assistance systems, and robotics. Before joining Virginia Tech, He was a Professor of Engineering and Applied Science at the George Washington University (GWU) and the Founding Director of the Center for Intelligent Systems Research from 1996 to 2015, the Director of the Transportation Safety and Security University Area of Excellence, from 2002 to 2015, and the Co-Founder of the National Crash Analysis Center in 1992 and its Director from 1998 to 2002 and 2013 to 2015. From 1989 to 1992, he was an Assistant Professor at Pennsylvania State University, York, PA, and an Engineer/Project Manager in the industry from 1983 to 1989. Dr. Eskandarian is a Fellow of ASME, a Fellow of IEEE (elevated in 2024), and a member of SAE professional societies. He received the SAE’s Vincent 2021 Bendix Automotive Electronics Engineering Award, the IEEE ITS Society’s Outstanding Researcher Award in 2017, and GWU’s School of Engineering Outstanding Researcher Award in 2013.
			\end{IEEEbiography}

\vfill

\end{document}